\documentclass{article} % For LaTeX2e
\usepackage{iclr2019_conference,times}

\usepackage{amsmath,amsfonts,bm}

\def\eqref#1{equation~\ref{#1}}
\def\1{\bm{1}}

\def\vtheta{{\bm{\theta}}}

\def\vx{{\bm{x}}}

\DeclareMathAlphabet{\mathsfit}{\encodingdefault}{\sfdefault}{m}{sl}
\SetMathAlphabet{\mathsfit}{bold}{\encodingdefault}{\sfdefault}{bx}{n}

\DeclareMathOperator*{\argmin}{arg\,min}

\usepackage{enumerate}
\usepackage{url}
\usepackage{amsmath}
\usepackage{amssymb}
\usepackage{bm}
\usepackage{hyperref}       % hyperlinks
\usepackage{url}            % simple URL typesetting
\usepackage{booktabs}       % professional-quality tables
\usepackage{amsfonts}       % blackboard math symbols
\usepackage{microtype}      % microtypography
\usepackage{amsmath}
\usepackage{amsthm}
\usepackage{enumerate}
\usepackage{comment}
\usepackage{url}
\usepackage{amsfonts}
\usepackage{mathrsfs}
\usepackage{amssymb}
\usepackage{bm}
\usepackage{algorithm}
\usepackage{algpseudocode}
\usepackage[normalem]{ulem} %for strikethrough
\usepackage{natbib}
\usepackage{graphicx}
\usepackage{geometry}
\usepackage{subfigure}
\usepackage{diagbox}
\usepackage[utf8]{inputenc}
\usepackage{float}
\usepackage{footnote}
\makeatletter
\renewcommand{\@thesubfigure}{\hskip\subfiglabelskip}
\makeatother

\theoremstyle{plain}

\theoremstyle{definition}

\title{RANDOM MASK: Towards Robust Convolutional Neural Networks}

\iclrfinalcopy
\author{
Tiange Luo$^{1}$\thanks{Equal contribution.},\; Tianle Cai$^1$\footnotemark[1],\; Mengxiao Zhang$^2$,\; Siyu Chen$^1$,\; Liwei Wang$^1$\\
$^1$Peking University, $^2$University of Southern California\\
\texttt{$^{1}$\{luotg,caitianle1998,siyuchen,wanglw\}@pku.edu.cn}, $^2$\texttt{zhan147@usc.edu} \\
}
\begin{document}

\maketitle

\begin{abstract}
%%Robustness of neural networks has been recently highlighted by the finding of \emph{adversarial examples}, e.g., well-designed inputs that are imperceptible for human but classified incorrectly by the network. In this paper, we introduce a simple but powerful technique \emph{Random Mask} to modify the existing CNN structure. Random Mask helps CNN to sift the features to be extracted by the distribution of the features captured by each filter. Then CNN with Random Mask will ignore irregular-distributed features and thus achieve robustness. In particular, CNN with Random Mask achieves the state-of-the-art performance with respect to black-box adversarial defense. Other experiments also show the robustness of CNN with Random Mask. Also, we use CNN with Random Mask to generate adversarial examples and find that there are actually some examples that can "fool" human within a small perturbation. This phenomenon both confirms that Random Mask really makes CNN to catch information more related to human perception and at the same time inspire us to rethink what is the robustness the network should really perform.
Robustness of neural networks has recently been highlighted by the \emph{adversarial examples}, i.e., inputs added with well-designed perturbations which are imperceptible to humans but can cause the network to give incorrect outputs. In this paper, we design a new CNN architecture that by itself has good robustness. We introduce a simple but powerful technique, \emph{Random Mask}, to modify existing CNN structures. We show that %Random Mask can significantly improve the robustness of CNN. In particular, 
CNNs with Random Mask achieve state-of-the-art performance against black-box adversarial attacks \emph{without} applying any adversarial training. %Moreover, we use CNN with Random Mask to generate adversarial examples and find that there are actually some slightly perturbed examples that can "fool" human. This phenomenon implies that Random Mask indeed makes CNN catch information that is more related to human perception. It also inspires us to rethink whether evaluating the robustness of neural networks by $\ell_\infty$-norm or other norm-based metric is appropriate.
We next investigate the adversarial examples which “fool” a CNN with Random Mask. Surprisingly, we find that these adversarial examples often “fool” humans as well. This raises fundamental questions on how to define adversarial examples and robustness properly.
%: What is a good definition of adversarial examples? How to define the robustness of a classifier? Is the widely-used $\ell_\infty$ norm a good metric for robustness?
\end{abstract}

\section{Introduction}
\label{sec:Introduction}
Deep learning~\citep{lecun2015deep}, especially deep Convolutional Neural Network (CNN)~\citep{lecun1998gradient}, has led to state-of-the-art results spanning many machine learning fields, such as image classification~\citep{he2016deep,hu2017squeeze,huang2017densely,simonyan2014very}, object detection~\citep{redmon2016you,girshick2015fast,ren2015faster}, image captioning ~\citep{vinyals2015show,xu2015show} and speech recognition~\citep{bengio2003neural,hinton2012deep}.

Despite the great success in numerous applications, recent studies have found that deep CNNs are vulnerable to some \emph{well-designed} input samples named as \emph{Adversarial Examples}~\citep{DBLP:journals/corr/SzegedyZSBEGF13}~\citep{biggio2013evasion}. Take the task of image classification as an example, for almost every commonly used well-performed CNN, attackers are able to construct a small perturbation on an input image to cause the model to give an incorrect output label. Meanwhile, the perturbation is almost \emph{imperceptible to humans}. % and should not influence the classification process by a human. 
Furthermore, these adversarial examples can easily \emph{transfer} among different kinds of CNN architectures~\citep{DBLP:journals/corr/PapernotMG16}.

%The reason why adversarial example is a serious problem is that the goal of learning usually requires robustness. 
%The fact that adversarial examples exist is a serious problem, since the goal of learning usually includes robustness. For example, empirically, people are not sensitive to small perturbation on images, so the image classifiers should be robust too. Just as ~\cite{Goodfellow2018Defense} suggests, both robustness and traditional supervised learning seem fully aligned. There are already lots of works considering the robustness of neural networks. \cite{DBLP:journals/corr/SzegedyZSBEGF13} used adversarial training to improve the robustness of neural networks by adding adversarial examples to the training data. Many other works focus on introducing randomness, including randomness in the input image~\citep{DBLP:journals/corr/abs-1711-01991} and randomness in the neural network structure~\citep{dhillon2018stochastic}. There are also a series of works focusing on performing some transformation to the input images to make the adversarial perturbation not sensitive to the neural networks~\citep{DBLP:journals/corr/abs-1711-00117}.

%The fact that adversarial examples exist is a serious problem, since the goal of learning usually includes robustness. For example, empirically, people are not sensitive to small perturbations on images, so the image classifiers should be robust too. 
Such adversarial examples raise serious concerns on deep neural network models as robustness is crucial in many applications. Just as ~\cite{Goodfellow2018Defense} suggests, both robustness and traditional supervised learning seem fully aligned. 
Recently, there is a rapidly growing body of work on this topic. %There are already lots of works considering the robustness of neural networks. 
One important line of research is adversarial training~\citep{DBLP:journals/corr/SzegedyZSBEGF13, madry2017towards, goodfellow2014, huang2015learning}. Although adversarial training gains some success, a major difficulty is that it tends to overfit to the method of adversarial example generation used at training time~\citep{buckman2018thermometer}. \cite{DBLP:journals/corr/abs-1711-01991} and \cite{DBLP:journals/corr/abs-1711-00117} propose defense methods by introducing randomness and applying transformations to the inputs respectively. \cite{dhillon2018stochastic} introduces random drop during the \emph{evaluation} of a neural network. However, ~\cite{DBLP:journals/corr/abs-1802-00420} contends that such transformation and randomness only provide a kind of “obfuscated gradient” and can be attacked by taking expectation over transformation (EOT) to get a meaningful gradient. \cite{papernot2016distillation} and \cite{katz2017reluplex} consider the non-linear functions in the networks and try to achieve robustness by adjusting them. There are also detection-based defense methods~\citep{metzen2017detecting, grosse2017statistical, meng2017magnet}, which add a process of detecting whether an input is adversarial.

In this paper, different from most of the existing methods, we take another approach to tackle the adversarial example problem. In particular, we aim to design a new CNN architecture which by itself enjoys robustness, without appealing to techniques such as adversarial training. To this end, we introduce Random Mask as a new ingredient of CNN. To be specific, we randomly select a set of neurons and remove them from the network \emph{before} training. Then the architecture of the network is \emph{fixed} during the training and testing process. %Note that Random Mask is different from dropout where neurons are randomly masked in each step \emph{during} training, and it can be applied very easily to common CNN structures such as ResNet with only a few changes of code. 
Note that Random Mask is different from dropout which randomly masks out neurons in each step \emph{during} training. 
%Random Mask is also different from SAP~\cite{dhillon2018stochastic} where neurons are masked by activation configuration during \emph{evaluation} phase which significantly reduces the accuracy on normal test data. 
In addition, Random Mask can be applied very easily to common CNN structures such as ResNet with only a few changes of code. We find that applying Random Mask to the shallow layers of the network is crucial for robustness. CNNs with properly designed Random Mask are far more robust than standard CNNs. In fact, our experimental results demonstrate that CNNs with Random Mask achieve state-of-the-art results against black-box attacks even when comparing with defense methods using adversarial training. Furthermore, CNNs with Random Mask maintain a high accuracy on normal test data, while low test accuracy is often regarded as a major weakness in many methods designed for achieving robustness.

\begin{figure}[htp]
\centering
\subfigure[Dog]{\includegraphics[width=0.4in]{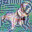}}
\hspace{0.05in}
\subfigure[Bird]{\includegraphics[width=0.4in]{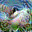}}
\hspace{0.05in}
\subfigure[Frog]{\includegraphics[width=0.4in]{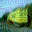}}
\hspace{0.05in}
\subfigure[Dog]{\includegraphics[width=0.4in]{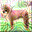}}
\hspace{0.05in}
\subfigure[Automobile]{\includegraphics[width=0.4in]{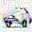}}
\hspace{0.05in}
\subfigure[Ship]{\includegraphics[width=0.4in]{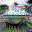}}
\hspace{0.05in}
\subfigure[Dog]{\includegraphics[width=0.4in]{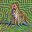}}
\hspace{0.05in}
\subfigure[Ship]{\includegraphics[width=0.4in]{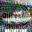}}
\hspace{0.05in}
\subfigure[Bird]{\includegraphics[width=0.4in]{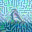}}
\hspace{0.05in}
\subfigure[Frog]{\includegraphics[width=0.4in]{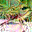}}
\hspace{0.05in}
\subfigure[Bird]{\includegraphics[width=0.4in]{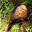}}
%\subfigure[Airplane]{\includegraphics[width=0.7in]{./img/2117_2_0.png}}
%\subfigure[Dog]{\includegraphics[width=0.7in]{./img/4857_4_5.png}}

\caption{Adversarial examples (generated by PGD against a network with Random Mask) that can “fool” a CNN with Random Mask. The labels here are the outputs of the network being “fooled”. The original images from CIFAR-10 and more examples can be found in Figure~\ref{cmp_ori} in Appendix~\ref{appendix: original images}.\label{cmp}} %ref
%\subfigure[Origin]{
%\begin{minipage}[b]{0.23\linewidth}
%\vspace{4pt}
%\includegraphics[width=1.3in]{./img/4997_7.png}
%\end{minipage}
%}
%\subfigure[ResNet18]{
%\begin{minipage}[b]{0.23\linewidth}
%\includegraphics[width=1.3in]{./img/15_8_9.png}\vspace{4pt}
%\includegraphics[width=1.3in]{./img/4929_7_3.png}
%\end{minipage}
%}
%\subfigure[RandomMask]{
%\begin{minipage}[b]{0.23\linewidth}
%\includegraphics[width=1.3in]{./img/15_8_0_1.png}\vspace{4pt}
%\includegraphics[width=1.3in]{./img/4997_7_5.png}
%\end{minipage}
%}
%\caption{\textbf{First row}:(a) The origin image is labeled as Ship. (b) The adversarial example generated by resnet18 and fooling the resnet18 as truck. (c) The adversarial example generated by our model and fooling the model as Plane. \textbf{Second Row}:
%(a) The origin image is labeled as Horse. (b) The adversarial example generated by resnet18 and fooling the resnet18 as cat. (c) The adversarial example generated by our model and fooling the model as Dog.}
\end{figure}

We next take a closer look at the adversarial examples generated particularly against CNNs with Random Mask. We investigate the adversarial examples that can “fool” our proposed architecture, i.e., the examples that are perturbed version of the original image, but are classified to a different label by the network. Surprisingly, we find that the adversarial examples which can “fool” a CNN with Random Mask often “fool” humans as well. It is difficult for humans to correctly classify these adversarial example, and in many cases humans make the same “incorrect” prediction as our network. Figure~\ref{cmp} shows a few adversarial examples generated by PGD (Basic iterative method)~\citep{kurakin2016adversarial} with respect to a CNN with Random Mask as well as the labels the network outputs. (Please also see Figure~\ref{cmp_ori} in Appendix~\ref{appendix: original images} for the original images and labels from CIFAR-10.) They are different from the typical adversarial examples generated against commonly used CNNs, which usually look like noisy versions of the original images and are easy to be correctly classified by humans.

These observations raise important questions: 1) How should we define adversarial examples? 2) How should we define robustness? Currently, an adversarial example is usually defined as a perturbed datum which lies in the neighborhood of the original data but has a different classification output by the network; and the robustness of a method is measured according to the proportion of these adversarial examples. However, if an adversarial example can also fool humans, it is more appropriate to say that the example does change the semantics of the datum to a certain extent. After all, why two images close to each other in terms of some distance (e.g., $\ell_\infty$) must belong to the same class? How close should they be so that they belong to the same class? Without complete answers to these questions, one should be very careful when measuring the robustness of a model merely according to currently-defined adversarial examples. Robustness is a subtle issue. We argue that one needs to rethink the robustness and adversarial examples from the definitions.

%Equipped with this perspective,
\newpage
Our main contributions are summarized as follows:

\begin{itemize}
\item We develop a very simple but effective method, \emph{Random Mask}. We show that combining with Random Mask, existing CNNs can be significantly more robust while maintaining high generalization performance. In fact, CNNs equipped with Random Mask achieve state-of-the-art performance against several black-box attacks, even when comparing with methods using adversarial training (See Table~\ref{Table 0}).
\item We investigate the adversarial examples generated against CNNs with Random Mask. We find that adversarial examples that can “fool” a CNN with Random Mask often fool humans as well. This observation requires us to rethink what are the right definitions of adversarial examples and robustness.
%\item We use a CNN with Random Mask to generate adversarial examples and find that relatively small perturbation of the origin image can actually change semantic information. We show that our adversarial examples (See Figure~\ref{cmp}) can even “fool” humans. These adversarial examples verify that CNN with Random Mask is more similar to the perception of human. Also, this finding motivates us to rethink whether the robustness with respect to $\ell_\infty$-norm or other norm-based measure is proper~\citep{madry2017towards}~\citep{xiao2018spatially} since we have found small but perceptible perturbation in the sense of $\ell_\infty$-norm.
%\item Experiments show that Random Mask really helps CNN to learn some extra information. In particular, the performance on randomly shuffled inputs (See Section~\ref{sec:Random Mask}, Appendix~\ref{appendix:random shuffle}) and the adversarial examples generated by CNN with Random Mask (See Figure~\ref{cmp}) reveal the great potential of applying Random Mask to lots of scenarios involved with machine learning such as object detection, 3D pose recognition, video tasks \textit{etc}.
\end{itemize}

\vspace{-0.2in}
\section{Random Mask}
\label{sec:Random Mask}
\begin{figure}[h]
\label{RandomMask}
\centering
\includegraphics[width=3in]{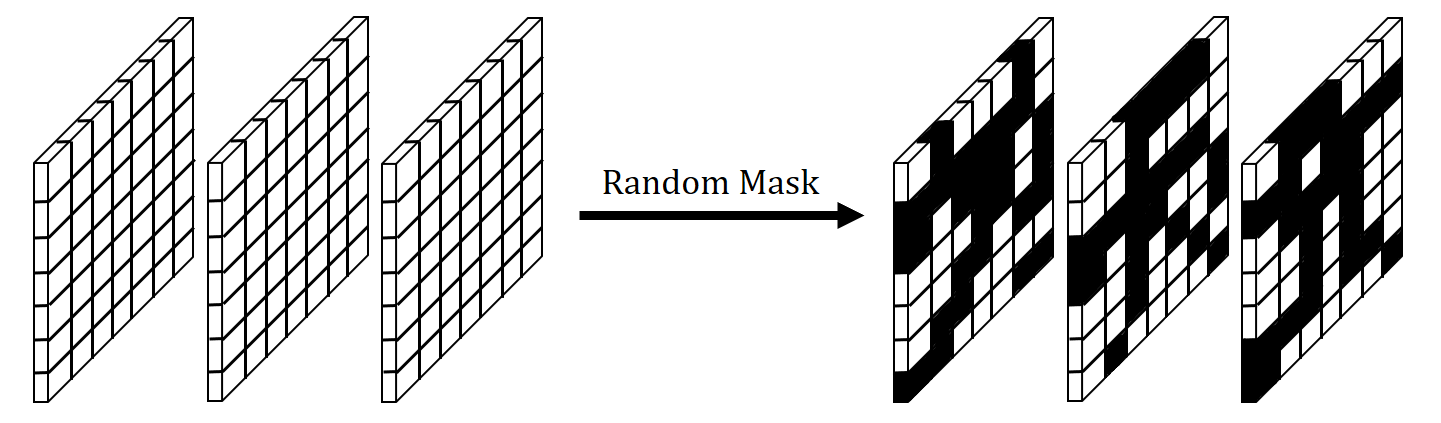}
\caption{An illustration of Random Mask applied to three \emph{channels} of a layer (neuron-wise). Note that the number of parameters in the network is \emph{not reduced} after applying Random Mask.}
\end{figure}

%%Just a draft
\vspace{-0.05in}
We propose Random Mask, a method to modify existing CNN structures. It randomly selects a set of neurons and removes them from the network \emph{before} training. Then the architecture of the network is \emph{fixed} during the training and testing process.
%Consider a CNN structure. The traditional CNN does convolution operation on each position uniformly. That is each neuron always catches information for an area of the feature map of former layer and for a convolution filter the area is of the same size. However, 
%Random Mask randomly masks some neurons of some selected layers \emph{before training}. It actually changes the network structure. Once we get a masked network, we can apply it to different tasks including training and testing without generating new Random Mask. In fact, experiments show that the performance of CNNs with differently generated Random Masks is quite similar. 
To apply Random Mask on a selected layer $Layer(j)$, suppose the input is $X_j$ and the output is $conv_j(X_j)\in \mathbb{R}^{m_j\times n_j\times c_j}$. We randomly generate a binary mask $mask(j) \in \{0,1\}^{m_j\times n_j\times c_j}$ by sampling uniformly within each channel. The \emph{drop rate} of the sampling process is called the \emph{ratio} (or \emph{drop ratio}) of Random Mask. Then we mask the neurons in position $(x,y,c)$ of the output of $Layer(j)$ if the $(x,y,c)$ element of $mask(j)$ is zero. More specifically, after Random Mask, we will not compute these masked neurons and make the next layer regard these neurons as having value zero during computation. A simple visualization of Random Mask is shown in Figure~\ref{RandomMask}. The Random Mask in fact decreases the computational cost in each epoch since there are fewer effective connections. Note that the \emph{number of parameters} in the convolutional kernels remains \emph{unchanged}, since we only mask neurons in the feature maps.

%\subsection{From Uniform Convolution to Random Mask}
\label{subsec:From uniform convolution to random mask}
\label{subsec:Information introduced by random mask}
In the standard setting, convolutional filter will be applied uniformly to every position of the feature map of the former layer. The success of this implementation is due to the reasonable assumption that if one feature is useful to be computed at some spatial position $(x,y)$, then it should also be useful to be computed at a different position $(x',y')$. Thus the original structure is powerful for feature extraction. Moreover, this structure leads to parameter sharing which makes the training process more efficient. However, the uniform application of filter also prevents the CNN from noticing the distribution of features. In other words, the network focuses on the \emph{existence} of a kind of feature but pays little attention to how this kind of feature \emph{distributes on the whole feature map} (of the former layer). Yet the pattern of feature distribution is important for humans to recognize and classify a photo, since empirically people would rely on some structured feature to perform classification. 

% Random shuffle experiment?
%By %%Experiment to be added
%we show this ignorance. 

%With Random Mask, each filter may only extract features from partial positions. Thus only features that distribute properly can be extracted by CNN with Random Mask. Just think of a toy example: when Random Mask for a filter masks all the neurons but one row in the channel, then if a kind of feature usually distributes in a row, it will not be caught by the masked filter with high probability. We do a straightforward experiments to verify our intuition. We train a CNN with Random Mask on the ImageNet and compare the accuracy for classifying random shuffled images (See Appendix~\ref{} for details), and find the CNN with Random Mask's accuracy consistently lower than that without Random Mask. This result shows CNN without Random Mask care more on whether a feature exists while CNN with Random Mask will limit the feature to be extracted. In Section~\ref{sec:Experiment}, we do extensive experiments to show that this asymmetric structure actually performances better than uniform convolution and bring robustness.

With Random Mask, each filter may only extract features from partial positions. More specifically, for one filter, only features which distribute consistently with the mask pattern can be extracted. Hence filters in a network with Random Mask may capture more information on the spatial structures of local features. Just think of a toy example: imagine Random Mask for a filter masks all the neurons but one row in the channel, if a kind of feature usually distributes in a column, it can not have strong response because the filter can only capture a small portion of the feature. 

We do a straightforward experiment to verify our intuition. We sample some images from ImageNet which can be correctly classified with high probability by both CNNs with and without Random Mask. We then randomly shuffle the images by patches, and compare the accuracy of classifying the shuffled images (See Appendix~\ref{appendix:random shuffle}). We find out that the accuracy of the CNN with Random Mask is consistently lower than that of normal CNN. This result shows that CNNs without Random Mask cares more about \emph{whether a feature exists} while CNNs with Random Mask will \emph{detect spatial structures} and \emph{limit poorly-organized features from being extracted}. 

We further explore how Random Mask plays its role in defending against adversarial examples. Recent observation~\citep{liu2018analyzing} of adversarial examples found that these examples usually change a patch of the original image so that the perturbed patch looks like a small part of the incorrectly classified object. This perturbed patch, although contains crucial features of the incorrectly classified object, usually appears at the wrong location and does not have the right spatial structure with other parts of the image. For example (See Figure 11 in~\cite{liu2018analyzing}), the adversarial example of a panda image is misclassified as a monkey because a patch of the panda skin is perturbed adversarially so that it alone looks like the monkey’s face. However, this patch does not form a right structure of a monkey with other parts of the images. By the properties of detecting spatial structures and limiting feature extraction, Random Mask can naturally help CNNs resist such adversarial perturbations.

\begin{figure}[htp]
\centering
\subfigure[Original Image]{\includegraphics[width=1in]{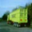}}
\hspace{0.2in}
\subfigure[Gaussian Noise]{\includegraphics[width=1in]{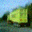}}
\hspace{0.2in}
\subfigure[Normal CNN]{\includegraphics[width=1in]{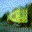}}
\hspace{0.2in}
\subfigure[Random Mask]{\includegraphics[width=1in]{./img/451_9_6.png}}
\caption{The first image is the original image, and the other three contain different types of small perturbations. Both the two adversarial examples on the right are predicted as frog by the corresponding models. However, only the image generated by the randomly masked CNN is capable of fooling humans. \label{cmpwithnorm}} 

\end{figure}

In complement to the observation we mentioned above, we also find that most adversarial perturbations generated against normal CNNs look like random noises which do not change the semantic information of the original image. In contrast, adversarial examples generated against CNNs with Random Mask tend to contain some well-organized features which sometimes change the classification results semantically (See Figure~\ref{cmpwithnorm} and Figure~\ref{cmp}). This phenomenon also supports our intuition that Random Mask helps to detect spatial structures and extract well-organized features via imposing limitations.

While the features that can be learned by each masked filter is limited, the \emph{randomness} helps us get plenty of diversified patterns. Our experiments show that these limited filters are enough for learning features. In other words, CNNs will maintain a high test accuracy after being applied with Random Mask. Besides, adding convolutional filters may help our CNN with Random Mask to increase test accuracy (See Section~\ref{subsec:On the Properties of Random Mask}). %%ref to Experiments 
Furthermore, our structure is naturally compatible to ensemble methods, and randomness makes ensemble more powerful (See Section~\ref{subsec:On the Properties of Random Mask}). %%ref to Experiments

However, it might not be appropriate to apply Random Mask to deep layers. The distribution of features is meaningful only when the location in feature map is \emph{highly related to the location in the original input image}, and the receptive field of each neuron in deep layers is too large. In Section~\ref{subsec:Shallow Layer Mask versus Deep Layer Mask}, there are empirical results which support our intuition.
%Notice that the distribution of features is meaningful only when the location in feature map is \emph{highly related to the location in the original input image}. We should not apply Random Mask to deep layers because the receptive field of each neuron of deep layers is too large. In Section~\ref{subsec:Shallow Layer Mask versus Deep Layer Mask}, there are empirical results which support our intuition.

%%ref to Appendix

\vspace{-0.15in}

\section{Experiments}
\label{sec:Experiment}
In this section, we provide extensive experimental analyses on the performance and properties of \emph{Random Mask} network structure. %The organization of this section is:
We first test the robustness of Random Mask (See Section~\ref{subsec:Adversarial defense via random mask}). Then we take a closer look at the adversarial examples that can “fool” our proposed architecture (See Section~\ref{subsec:Adversarial Examples Question Mark}). After that we explore properties of Random Mask, including where and how to apply Random Mask, by a series of comparative experiments (See Section~\ref{subsec:Shallow Layer Mask versus Deep Layer Mask}). Some settings used in our experiments are listed below:

%\textbf{Dataset} We mainly focus on image classification tasks and choose three different datasets: MNIST, CIFAR-10. %and tiny-ImageNet. %and tiny-ImageNet~\citep{vinyals2016matching}.
%Tiny-ImageNet consists of 60,000 color images of size $84\times84$ with 100 classes, each having 600 examples. It is a more complex dataset than CIFAR-10. 
%As the performances on these datasets are similar, we only show the result of CIFAR-10 in this section and leave other experiment results in the Appendix~\ref{appendix:More Experiment Result}.%%ref to Appendix

\textbf{Network Structure.} We apply Random Mask to several target networks, including ResNet-18~\citep{he2016deep}, ResNet-50, DenseNet-121~\citep{huang2017densely}, SENet-18~\citep{hu2017squeeze} and VGG-19~\citep{simonyan2014very}. The effects of Random Mask on those network structures are quite consistent. For brevity, we only show the defense performance on ResNet-18 in the main body and leave more experimental results in the Appendix~\ref{appendix:More Experiment Result}. The 5-block structure of ResNet-18 is shown in the Appendix~\ref{ResNet18}. The blocks are labeled \(0,1,2,3,4\) and the $0^{\text{th}}$ block is the first convolution layer. 
%For simplicity, we regard each of the five blocks as a whole.
We divide these five blocks into two parts - the relative shallow ones (the $0^{\text{th}},1^{\text{st}},2^{\text{nd}}$ blocks) and the deep ones (the $3^{\text{rd}},4^{\text{th}}$ blocks). For simplicity, we would like to regard each of these two parts as a whole in this section to avoid being trapped by details. We use “$\sigma$-Shallow” and “$\sigma$-Deep” to denote that we apply Random Mask with drop ratio \(\sigma\) to the shallow blocks and to the deep blocks in ResNet-18 respectively.

\textbf{Attack Framework.} The \emph{accuracy under black-box attack} serves as a common criterion of robustness. We will use it when selecting model parameters and comparing Random Mask to other similar structures. To be more specific, 
%by using the methods mentioned in Section~\ref{subsec:Information introduced by random mask}, 
by using FGSM ~\citep{goodfellow2014}, PGD ~\citep{kurakin2016adversarial} with \(\ell_\infty\) norm and CW attack ~\citep{DBLP:journals/corr/CarliniW16a} with \(\ell_2\) norm (See Appendix~\ref{appendix:attack approaches} for details on these attack approaches),
%% Whether put the method in the main part need discussion
 we generate adversarial examples against different neural networks. The performances on adversarial examples generated against different networks are quite consistent. For brevity, we only show the defense performance against part of the adversarial examples generated by using DenseNet-121 on dataset CIFAR-10 in this section, and leave more experimental results obtained by using other adversarial examples in the Appendix~\ref{appendix:More Experiment Result}. We use \(\textbf{FGSM}_{16},\textbf{PGD}_{16}, \textbf{PGD}_{32},\textbf{CW}_{40}\) to denote attack method FGSM with step size \(\epsilon=16\), PGD with perturbation scale \(\alpha=16\) and step number \(20\), PGD with perturbation scale \(\alpha=32\) and step number \(40\), CW attack with confidence \(\kappa=40\) respectively. The step size of both PGD methods are selected to be \(\epsilon=1\). We would like to point out that these attacks are really powerful that a normal network cannot resist these attacks.

%In Section~\ref{subsec:Adversarial defense via random mask}, we have suggested that

%For presentational clearness we will only show the success ratio under black box attack generated by DenseNet-121 with PGD.
% More complete data of other attack generated by other networks with different methods and strengths are shown in Appendix %%ref

%\subsection{Model selection}
%\label{subsec:Model selection}
%We first do experiments to evaluate the performance and select a specific strategy to apply Random Mask for further experiments. We deploy several experiments to compare whether we should apply Random Mask in shallow layer or in deep layer and what mask ratio should we choose to weigh the generalization and robustness. 

%%After applying Random Mask to some of its layers, we retrain the network and see its test accuracy and the successful rate of attack from the adversarial examples generated by other neural networks.
\vspace{-0.05in}
\subsection{Robustness via Random Mask}
\label{subsec:Adversarial defense via random mask}

Random Mask is not specially designed for adversarial defense, but as Random Mask introduces information that is essential for classifying correctly, it also brings robustness. As mentioned in Section~\ref{subsec:From uniform convolution to random mask}, normal CNN structures may allow adversary to \emph{inject features imperceptible to humans} into images that can be recognized by CNN. %Then these extracted features may fool CNN.
Yet Random Mask limits the process of feature extraction, so noisy features are less likely to be preserved.%when they are actually not well-organized.
\vspace{-0.05in}
\subsubsection{Robustness to Black-box Attack} 

%more kinds of perturbation such as gaussian noise
The results of our experiments %show more properties of Random Mask and also 
show the strengths of applying Random Mask to adversarial defense. In fact, Random Mask can help existing CNNs reach state-of-the-art performance against the black-box attacks we use (See Table~\ref{Table 0}). In Section~\ref{subsec:On the Properties of Random Mask}, we will provide more experimental results to show that this asymmetric structure performs better than normal convolution and enhances robustness. %%Add Table

\begin{table}[h]
\begin{center}
\begin{tabular}{cccc}
\toprule
\bf Model  &\bf FGSM &\bf PGD
&\bf Test Accuracy
\\ \midrule
Normal ResNet-18 &26.99\%	&7.56\%	&95.33\%
\\
Vanilla (Madry) &85.60\% &86.00\%	&87.30\%
\\
Random Mask &\bf 86.31\%	& \bf 90.30\% & \bf 90.08\%
%\\
%Random Mask (ensemble) &\bf88.56\%	&\bf 91.28\% &91.77\%
\\ \bottomrule
\end{tabular}
\end{center}
%\caption{Performance of black-box defense under the setting of \cite{madry2017towards} (See Appendix~\ref{appendixsubsec:Madry Setting} for the complete setting). We use model under adversarial training in ~\cite{madry2017towards} as a vanilla model. It is regarded as a state-of-the-art adversarial defense method. Our model is only trained on \emph{clean data}. The ratio of Random Mask here is selected to balance the performance of robustness and generalization. See Figure~\ref{ratio} in Appendix~\ref{appendix:ratio} for complete results on the performance of Random Mask with different ratios.\label{Table 0}}

\caption{Performance of black-box defense under the setting of \cite{madry2017towards} (See Appendix~\ref{appendixsubsec:Madry Setting} for the complete setting). We use model under adversarial training in ~\cite{madry2017towards} as a vanilla model. It is regarded as a state-of-the-art adversarial defense method. Our model is only trained on \emph{clean data}. The ratio of Random Mask here is selected to balance the performance of robustness and generalization. See Figure~\ref{ratio} in Appendix~\ref{appendixsubsec:Madry Setting} for results on the performance of Random Mask with different ratios.\label{Table 0}}
\end{table}

%\subsubsection{Mask Ratio Comparison}
%\label{subsec:Mask Ratio Comparison}
%Next, We fix shallow layers to apply Random Mask and evaluate the influence of Random Mask ratio.

%The results show that as the ratio of mask increase, the test accuracy decrease and the success ratio against black box attack increase. This show the trade-off between catching information of more position and sift information via relative location more rigorously. Although adjusting the global network structure can help to maintain the test accuracy, we would not do so because the origin global structure with Random Mask is good enough to be compared with the one without Random Mask. Balancing these two aspects, we would like to choose the network structures with \((\{1,2\},0.5), (\{1,2\},0.9)\)
%as two typical examples for further experiments.

\subsubsection{Robustness to Random Noise}
\label{subsec:On the Robustness of Random Mask}
%To see the advantage that Random Mask has, we do a series of systematic experiments to compare the robustness of Random Mask with that of several similar structures from different aspects.
%Besides the power of Random Mask to improve CNN's robustness against \emph{black-box attacks}. Table~\ref{Table 0} in Section~\ref{sec:Random Mask} and Appendix~\ref{appendix:More Experiment Result} %ref
%also show that with proper parameters, Random Mask can give \emph{state-of-the-art} black-box defense performance.
%%Table to be added
%%{1,4,2},{1,6,4},{1,10,8}

\begin{figure}[h]
\centering
\includegraphics[width=2.75in]{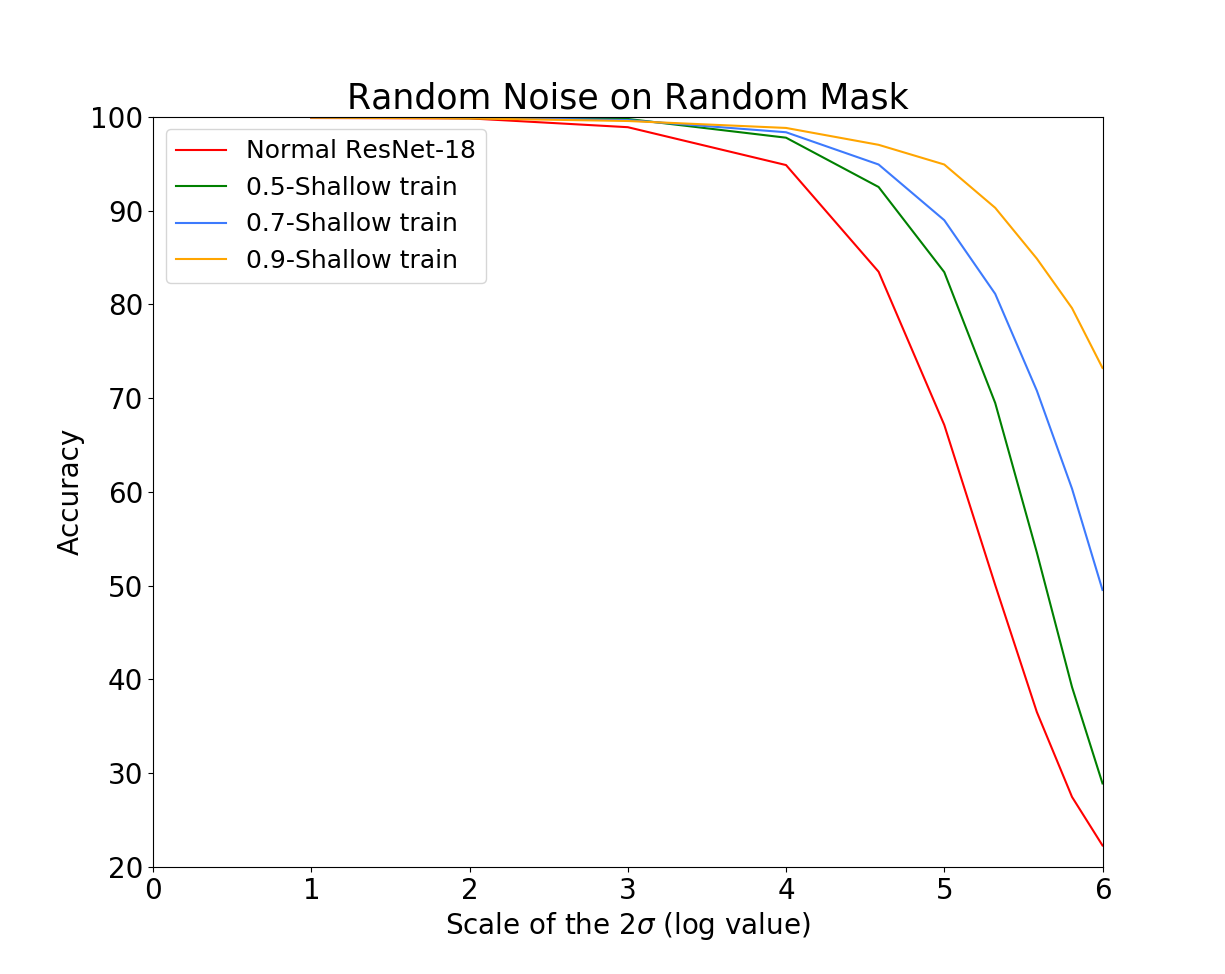}
\caption{Input images with random Gaussian noises, Random Mask versus normal network. The network with Random Mask has far better robustness than the original network.\label{RandomNoise}}
\end{figure}

%\subsubsection{Robustness to Random Noise}
%In Section ~\ref{subsec:Shallow Layer Mask versus Deep Layer Mask} and ~\ref{subsec:On the Properties of Random Mask}, we show that our Random Mask outperforms original network structure in defending black-box attack with little test accuracy cost. However, n
%In this subsection, we would like to further evaluate the robustness of CNN with Random Mask. We mainly focus on and 
Beside the robustness against black-box attack, we also evaluate the robustness to random noise. Note that although traditional network structures are vulnerable to adversarial examples, they are still robust to the images perturbed with small Gaussian noises. To see whether our structure also enjoys such property, or even has better robustness in this sense, we feed input images with random Gaussian noises to networks with Random Mask. More specifically, in order to obtain noises of scales similar to the adversarial perturbations, we generate i.i.d. Gaussian random variables \(x\sim N(0, \sigma^2)\), where \(\sigma\in\{1,2,4,8,12,16,20,24,28,32\}\), clip them to the range \([-2\sigma,2\sigma]\) and then add them to every pixel of the input image. % and clip the image to the \(2\sigma-l_\infty\) ball of the original one. 
The results of the experiments are shown in Figure~\ref{RandomNoise}. We can see that networks with Random Mask always have higher accuracy than a normal network.

%From the Figure~\ref{RandomNoise}, we can see that by applying Random Mask, our network still enjoys the stability to small random Gaussian noise. Moreover, for larger random noise, Random Mask shows far better robustness than the original network. This means that compared with the traditional structure, our Random Mask can not only defend black-box adversarial examples well, but also be more stable with random Gaussian noise.
\vspace{-0.1in}
\subsection{Adversarial Examples?}
\label{subsec:Adversarial Examples Question Mark}
We evaluate the performance of CNNs with Random Mask under white-box attack (See Appendix~\ref{Appendix: white}). With neither obfuscated gradient nor gradient masking, Random Mask can still improve defense performance under various kinds of white-box attack. Also, by checking adversarial images that are misclassified by our network, we find most of them have vague edges and \emph{can hardly be recognized by humans}. 
%In contrast, adversarial examples generated by most commonly-used networks are more like just adding some noise. 
This result coincides with the theoretical analysis in \cite{shafahi2018adversarial, fawzi2018adversarial} that real adversarial examples may be inevitable in some way. 
% See Appendix~\ref{appendix: random selected} for a randomly selected set of them. 
In contrast, adversarial examples generated against normal CNNs are more like simply adding some non-sense noise which can be ignored by human. This phenomenon also demonstrates that Random Mask really helps networks to catch more information related to real human perception. Moreover, just as Figure~\ref{cmp} shows, with the help of Random Mask, we are able to find small perturbations that can actually change the semantic meaning of images for humans. So should we still call them “adversarial examples”? How can we get more reasonable definitions of adversarial examples and robustness? These questions seem severe due to our findings.

\vspace{-0.1in}
\subsection{Properties of Random Mask}
\label{subsec:On the Properties of Random Mask}
We then show some properties of Random Mask including the appropriate positions to apply Random Mask, the benefit of breaking symmetry, the diversity introduced by randomness and the extensibility of Random Mask via structure adjustment and ensemble methods. We conduct a series of comparative experiments and we will continue to use black-box defense performance as a criterion of robustness. For brevity, we only present the results of a subset of our experiments in Table~\ref{Table 3}. Full information on all the experiments can be found in Appendix~\ref{Appendix: resnet18 experiments}.

\begin{table}[h]
\begin{center}
\begin{tabular}{crrrrc}
\toprule
\multicolumn{1}{c}{\bf Network Structure} &\bf $\text{FGSM}_{16}$ &\bf $\text{PGD}_{16}$ &\bf $\text{PGD}_{32}$ &\bf 
$\text{CW}_{40}$
&\bf Test Accuracy\\
\midrule
Normal ResNet-18 &14.91\%	&2.96\%	&2.26\% &8.23\% &95.33\%
\\ \hline
\specialrule{0em}{1pt}{1pt}
\multicolumn{1}{l}{0.3-Shallow} &23.29\%	&14.53\%	&5.73\% &36.95\% &94.03\%
\\
\multicolumn{1}{l}{0.5-Shallow} &30.86\%	&26.50\%	&10.33\% &54.02\%&93.39\%
\\
\multicolumn{1}{l}{0.7-Shallow}  &48.57\%   &47.76\% &21.39\% &73.70\% &91.83\%
\\
\multicolumn{1}{r}{0.3-Deep} &14.62\%	&1.95\%	&1.30\% &7.88\%&95.16\%
\\
\multicolumn{1}{r}{0.5-Deep}  &10.76\%	&2.57\%	&4.52\% &7.19\%&94.94\%
\\
\multicolumn{1}{r}{0.7-Deep}  &11.23\%	&3.24\%	&2.64\% &10.10\%&94.61\%
\\
0.3-Shallow, 0.3-Deep &24.15\%	&12.67\%	&6.75\% &29.65\%&94.16\%
\\
0.3-Shallow, 0.7-Deep &11.26\% &7.94\% &5.77\% &23.31\%&93.44\%
\\
0.7-Shallow, 0.7-Deep &27.43\%	&32.72\%	&16.26\% &62.47\%&89.78\%
\\
0.7-Shallow, 0.3-Deep &40.58\% &42.95\% &19.00\% &68.58\%&91.23\%

\\ \hline
\specialrule{0em}{1pt}{1pt}
0.5-Shallow &30.86\%	&26.50\%	&10.33\% &54.02\%&93.39\%
\\
%0.7-Shallow  &48.77\%   &47.76\% &21.39\% &73.70\% &91.83\%
%\\
0.9-Shallow  &79.93\%  &83.08\% &55.02\% &89.67\% &87.68\%
\\ \hline
\specialrule{0em}{1pt}{1pt}
$\text{0.5-Shallow}_{\text{DC}}$ &12.15\%	&4.68\%	&4.05\% &12.72\% &94.97\%
\\
%$\text{0.7-Shallow}_{\text{DC}}$&17.82\%	&7.51\%	&4.52\% &19.34\% &94.23\%
%\\
$\text{0.9-Shallow}_{\text{DC}}$ &19.00\% &19.33\% &10.08\% &44.80\% &93.27\%
\\ \hline
\specialrule{0em}{1pt}{1pt}
$\text{0.5-Shallow}_{\text{SM}}$ &48.86\%	&44.04\%	&19.81\%&72.07\% &92.57\%
\\
%$\text{0.7-Shallow}_{\text{SM}}$&48.06\%	&57.62\%	&24.81\%&79.55\% &89.81\%
%\\
$\text{0.9-Shallow}_{\text{SM}}$ &39.40\% &50.40\% &29.23\% &65.38\% &74.28\%
\\ \hline
\specialrule{0em}{1pt}{1pt}
$\text{0.5-Shallow}_{\times 2}$&20.78\% &12.51\%  &5.38\% &34.00\% &94.12\%
\\
%$\text{0.7-Shallow}_{\times 2}$&31.08\%	&29.83\% &11.44\% &60.17\% &93.01\%
%\\
$\text{0.9-Shallow}_{\times 2} $&68.83\% &66.86\% &37.51\% &82.74\% &90.49\%
\\
$\text{0.9-Shallow}_{\times 4} $&59.64\% &59.15\% &32.29\% &78.88\% &90.57\%
\\ \hline
\specialrule{0em}{1.5pt}{1.5pt}
$\text{Normal ResNet-18}_{\text{EN}}$ &16.24\% &2.22\%  &1.46\%  &8.58\% &96.12\%
\\
$\text{0.5-Shallow}_{\times 2, \text{EN}}$ &19.84\% &11.86\%  &5.30\% &37.37\% &95.24\%
\\
$\text{0.5-Shallow}_{\text{EN}}$ &31.38\% &27.58\% &9.97\%  &58.07\% &94.56\%
\\
$\text{0.9-Shallow}_{\text{EN}}          $&81.95\% &85.14\% &56.02\% &91.36\% &89.45\%
\\
\bottomrule
\end{tabular}
\end{center}
%Extensive experiments about Random Mask.
\caption{A subset of our experiments presented in Appendix~\ref{Appendix: resnet18 experiments} to show properties of Random Mask. $\sigma\text{-Shallow}_{\text{DC}}$, $\sigma\text{-Shallow}_{\text{SM}}$, $\sigma\text{-Shallow}_{\times n}$ and $\sigma\text{-Shallow}_{\text{EN}}$ mean dropping channels with ratio $\sigma$, applying same mask with ratio $\sigma$, increasing channel number to $n$ times with mask ratio $\sigma$ for every channel and ensemble five models with different masks of same ratio $\sigma$ respectively. The entries in the middle four columns are success rates of defense under different settings. This is also .\label{Table 3}}

\end{table}

\textbf{Masking Shallow Layers versus Masking Deep Layers.}
\label{subsec:Shallow Layer Mask versus Deep Layer Mask}
In the last paragraph of Section~\ref{subsec:Information introduced by random mask} , we give an intuition that deep layers in a network should not be masked. To verify this, we do extensive experiments on ResNet-18 with Random Mask applied to different parts.
%% and adversarial examples generated from different trained neural network. 
%%Appendix 4-block structure of ResNet-18
We apply Random Mask with different ratios on the shallow blocks and on the deep blocks respectively. %The results are listed in the following table.
%%\begin{table}[h]
%%\begin{center}
%%\begin{tabular}{lrrrrc}
%%\toprule
%%\multicolumn{1}{c}{\bf Network Structure} &\bf $\text{FGSM}_{16}$ &\bf $\text{PGD}_{16}$ &\bf $\text{PGD}_{32}$ &\bf 
%%$\text{CW}_{40}$
%%&\bf Test Accuracy\\
%%\midrule
%%
%%
%%\\ \bottomrule
%%\end{tabular}
%%\end{center}
%%\caption{Masking Shallow Blocks versus Masking Deep Blocks\label{Table 1}
%%}
%%\end{table}
%%(Experiment results not updated, PGD scale) Note that %ref
%%show PGD is a universal first order adversary, in other words, developing robustness against PGD attacks also implies resistance against many other first order attacks. Therefore, for the sake of brevity, we only show the results with PGD adversarial examples and CIFAR-10 dataset. We leave more experiment results in the Appendix. To generate adversarial examples, We use PGD with ...... on each well-trained network and test them on the networks with Random Mask at different layers. From Table~\ref{Table 2}, we can actually see that as Rand
Results in Table~\ref{Table 3} accord closely with our intuition. Comparing the success rate of black-box attacks on the model with the same drop ratio but different parts being masked, we find that applying Random Mask to \emph{shallow layers} enjoys significantly lower adversarial attack success rates. This verifies that shallow layers play a more important role in limiting feature extraction than the deep layers. Moreover, only applying Random Mask on shallow blocks can achieve better performance than applying Random Mask on both shallow and deep blocks, which also verifies our intuition that dropping elements with large receptive fields is not beneficial for the network. In addition, we would like to point out that ResNet-18 with Random Mask significantly \emph{outperforms} the normal network in terms of \emph{robustness}.

\textbf{Random Mask versus Channel Mask.}\label{subsec:Random Mask versus Channel Mask} As our Random Mask applies independent random masks to different channels in a layer, we actually break the symmetry of the original CNN structure. To see whether this asymmetric structure would help, we try to directly drop whole channels instead of neurons using the same drop ratio as the Random Mask and train it to see the performance. This channel mask does not hurt the symmetry while also leading to the same decrease in convolutional operations. Table~\ref{Table 3} shows that although our Random Mask network suffers a small drop in test accuracy due to the high drop ratio, we have a great gain in the robustness, compared with the channel-masking network. %Instead of carelessly capturing features of input images, our Random Mask limits the distribution of features to be extracted. However, both the baseline network and the channel-dropping network still keep the symmetry to the input image and thus suffer a high black-box attack success rate.

\textbf{Random Mask versus Same Mask.}
\label{subsec:Random Mask versus Same Mask} The randomness in generating masks in different channels and layers allows each convolutional filter to focus on different patterns of feature distribution. We show the essentialness of generating various masks per layer via experiments that compare Random Mask to a method that only randomly generates one mask per layer and uses it in every channel. Table~\ref{Table 3} shows that applying the same mask to each channel will decrease the test accuracy. This may result from the limitation of expressivity due to the monotone masks at every masked layer. In fact, we can illustrate such limitation using simple calculations. Since the filters in our base network ResNet-18 is of size $3\times 3$, each element of the feature maps after the first convolutional layer can extract features from at most $9$ pixels in the original image. This means that if we use the same mask and the drop ratio is \(90\%\), only at most \(9\times10\%\) of the input image can be caught by the convolutional layer, which would cause severe loss of input information.
%The success ratio of same mask is usually a little higher than random mask, this result if due to similar reason as increase mask ratio, that the rigorousness for features to locate at some specific positions may be benefit for the robustness.

%% \subsubsection{Random Mask versus Dropout Network after Training}
%% \label{subsec:Random Mask versus Dropout Network after Training}
%% (Ask for suggestion) The Random Mask is applied before training, which is totally different from dropout or SAP ~\citep{dhillon2018stochastic}. Meanwhile, we do not use any prior of adversarial attack while applying Random Mask. To see the benefit, we compare the Random Mask to a network that do dropout on a trained network for defense adversarial attack. %%specific structure
%% We find that although we carefully scale the weight after dropout, the decrease of test accuracy is so large that when we dropout 50\% and 90\%, the test accuracy are ...%%
%% respectively.

%\subsubsection{Improvements for Random Mask}
%\label{subsec:Improve Random Mask by Changing the Network Structure}
\textbf{Increase the Number of Channels.} %As Random Mask introduces more pattern and location knowledge than the origin network. 
%For balancing the cost of masking many neurons in each channel, 
In order to compensate the loss of masking many neurons in each channel, it is reasonable that we may need more convolutional filters for feature extraction. Therefore, we try to increase the number of channels at masked layers. Table~\ref{Table 3}  shows that despite ResNet-18 is a well-designed network structure, increasing channels does help the network with Random Mask to get higher test accuracy while maintaining good robustness performance. 

%\begin{table}[h]
%\label{Table 5}
%\begin{center}\begin{tabular}{crrrrc}
%\toprule
%\multicolumn{1}{c}{\bf Network Structure} &\bf $\text{FGSM}_{16}$ &\bf $\text{PGD}_{16}$ &\bf $\text{PGD}_{32}$ &\bf 
%$\text{CW}_{50}$
%&\bf Test Accuracy\\
%\midrule
%0.5-Shallow &31.06\%	&26.50\%	&10.33\% &54.02\%&93.39\%
%\\
%0.7-Shallow  &48.77\%   &47.76\% &21.39\% &73.70\% &91.83\%
%\\
%0.9-Shallow  &79.97\%  &83.08\% &55.02\% &89.67\% &87.68\%
%\\
%
%\\ \bottomrule
%\end{tabular}
%\end{center}
%\caption{Random Mask with more channels. $\sigma\text{-Shallow}_{\times n}$ means .}
%\end{table}

%At the same time, network structure can also be adjusted to inspire the potential of Random Mask. Since our aim is just to introduce the Random Mask, we do few experiments of finding the more powerful structure, which is already enough for showing the potential of Random Mask. We would like to underline that the ResNet-18 itself is well-designed thus the partial adjustment on it may be less strong.

\textbf{Ensemble Methods.} Thanks to the diversity of Random Mask, we may directly use several networks with the same structure but different Random Masks and ensemble them. Table~\ref{Table 3} shows that such ensemble methods can improve a network with Random Mask in both test accuracy and robustness.

% \begin{table}[h]
% \label{Table 6}
% \begin{center}\begin{tabular}{crrrrc}
% \toprule
% \multicolumn{1}{c}{\bf Network Structure} &\bf $\text{FGSM}_{16}$ &\bf $\text{PGD}_{16}$ &\bf $\text{PGD}_{32}$ &\bf 
% $\text{CW}_{50}$
% &\bf Test Accuracy\\
% \midrule
% \\0.5-Shallow &9.27\%	&5.80\%	&x\% &94.35\%
% \\
% 0.7-Shallow &13.18\%	&6.92\%	&x\% &92.48\%
% \\
% 0.9-Shallow  &22.08\% &10.56\% &x\% &92.80\%
% \\
% $\text{0.5-Shallow}_{\times 2}$&20.91\%	&4.66\%	&x\% &x\% &93.77\%
% \\
% $\text{0.7-Shallow}_{\times 2}$&4.49\%	&3.14\%	&x\% &x\% &93.48\%
% \\
% $\text{0.9-Shallow}_{\times 2}$&20.35\% &9.80\% &x\% &x\% &91.82\%
% \\ \bottomrule
% \end{tabular}
% \end{center}
% \caption{Apply ensemble methods on networks with Random Mask.}
% \end{table}

\section{Conclusion and future directions}
%%Video, Object detection...
%In conclusion, we introduce and experiment on Random Mask, a modification on existing CNNs that makes CNNs capture more information including the pattern of feature distribution. Consequently CNNs with Random Mask can give more careful inference and achieve better robustness. It is worth mentioning that Random Mask helps CNNs reach state-of-the-art performance in black-box defense. In addition to robustness, since Random Mask captures extra information that is highly related to human perception and limits feature extraction (See Section~\ref{sec:Random Mask}, Appendix~\ref{appendix:random shuffle}), it may have further applications in lots of other scenarios involved with deep learning where we do not only care about whether a feature exists, such as object detection, 3D pose recognition, video tasks \textit{etc}. Futhermore, from the adversarial examples generated against CNNs with Random Mask, we propose a question of whether the popular criterion of robustness using norm-based metric is really appropriate. We hope our work can inspire more people to rethink adversarial examples and the robustness of neural networks.

%In this paper, we introduce Random Mask, a modification on existing CNNs that makes them capture more information including the image patterns in different locations. 
In conclusion, we introduce and experiment on Random Mask, a modification of existing CNNs that makes CNNs capture more information including the pattern of feature distribution. We show that CNNs with Random Mask can achieve much better robustness while maintaining high test accuracy. More specifically, by using Random Mask, we reach state-of-the-art performance in several black-box defense settings. Another insight resulting from our experiments is that the adversarial examples generated against CNNs with Random Mask actually change the semantic information of images and can even “fool” humans. %This finding inspires us to rethink whether the popular criterion of robustness using norm-based metric is reasonable. 
We hope that this finding can inspire more people to rethink adversarial examples and the robustness of neural networks. %Besides robustness, as Random Mask captures more information in relative margins and features, we argue that our framework can be applied to a wide range of machine learning and computer vision tasks, such as object detection, 3D pose recognition, video tasks, \textit{etc}.  
\begin{comment}

\end{comment}

\bibliography{iclr2019_conference}

\begin{thebibliography}{36}
\providecommand{\natexlab}[1]{#1}
\providecommand{\url}[1]{\texttt{#1}}
\expandafter\ifx\csname urlstyle\endcsname\relax
  \providecommand{\doi}[1]{doi: #1}\else
  \providecommand{\doi}{doi: \begingroup \urlstyle{rm}\Url}\fi

\bibitem[Athalye et~al.(2018)Athalye, Carlini, and
  Wagner]{DBLP:journals/corr/abs-1802-00420}
Anish Athalye, Nicholas Carlini, and David~A. Wagner.
\newblock Obfuscated gradients give a false sense of security: Circumventing
  defenses to adversarial examples.
\newblock \emph{CoRR}, abs/1802.00420, 2018.
\newblock URL \url{http://arxiv.org/abs/1802.00420}.

\bibitem[Bengio et~al.(2003)Bengio, Ducharme, Vincent, and
  Jauvin]{bengio2003neural}
Yoshua Bengio, R{\'e}jean Ducharme, Pascal Vincent, and Christian Jauvin.
\newblock A neural probabilistic language model.
\newblock \emph{Journal of machine learning research}, 3\penalty0
  (Feb):\penalty0 1137--1155, 2003.

\bibitem[Biggio et~al.(2013)Biggio, Corona, Maiorca, Nelson, {\v{S}}rndi{\'c},
  Laskov, Giacinto, and Roli]{biggio2013evasion}
Battista Biggio, Igino Corona, Davide Maiorca, Blaine Nelson, Nedim
  {\v{S}}rndi{\'c}, Pavel Laskov, Giorgio Giacinto, and Fabio Roli.
\newblock Evasion attacks against machine learning at test time.
\newblock In \emph{Joint European conference on machine learning and knowledge
  discovery in databases}, pp.\  387--402. Springer, 2013.

\bibitem[Buckman et~al.(2018)Buckman, Roy, Raffel, and
  Goodfellow]{buckman2018thermometer}
Jacob Buckman, Aurko Roy, Colin Raffel, and Ian Goodfellow.
\newblock Thermometer encoding: One hot way to resist adversarial examples.
\newblock In \emph{International Conference on Learning Representations}, 2018.

\bibitem[Carlini \& Wagner(2016)Carlini and
  Wagner]{DBLP:journals/corr/CarliniW16a}
Nicholas Carlini and David~A. Wagner.
\newblock Towards evaluating the robustness of neural networks.
\newblock \emph{CoRR}, abs/1608.04644, 2016.
\newblock URL \url{http://arxiv.org/abs/1608.04644}.

\bibitem[Dhillon et~al.(2018)Dhillon, Azizzadenesheli, Lipton, Bernstein,
  Kossaifi, Khanna, and Anandkumar]{dhillon2018stochastic}
Guneet~S Dhillon, Kamyar Azizzadenesheli, Zachary~C Lipton, Jeremy Bernstein,
  Jean Kossaifi, Aran Khanna, and Anima Anandkumar.
\newblock Stochastic activation pruning for robust adversarial defense.
\newblock \emph{arXiv preprint arXiv:1803.01442}, 2018.

\bibitem[Fawzi et~al.(2018)Fawzi, Fawzi, and Fawzi]{fawzi2018adversarial}
Alhussein Fawzi, Hamza Fawzi, and Omar Fawzi.
\newblock Adversarial vulnerability for any classifier.
\newblock \emph{arXiv preprint arXiv:1802.08686}, 2018.

\bibitem[Girshick(2015)]{girshick2015fast}
Ross Girshick.
\newblock Fast r-cnn.
\newblock In \emph{Proceedings of the IEEE international conference on computer
  vision}, pp.\  1440--1448, 2015.

\bibitem[Goodfellow(2018)]{Goodfellow2018Defense}
Ian Goodfellow.
\newblock Defense against the dark arts: An overview of adversarial example
  security research and future research directions.
\newblock \emph{arXiv preprint arXiv:1806.04169}, 2018.

\bibitem[Goodfellow et~al.(2015)Goodfellow, Shlens, and
  Szegedy]{goodfellow2014}
Ian Goodfellow, Jonathon Shlens, and Christian Szegedy.
\newblock Explaining and harnessing adversarial examples.
\newblock In \emph{International Conference on Learning Representations}, 2015.
\newblock URL \url{http://arxiv.org/abs/1412.6572}.

\bibitem[Grosse et~al.(2017)Grosse, Manoharan, Papernot, Backes, and
  McDaniel]{grosse2017statistical}
Kathrin Grosse, Praveen Manoharan, Nicolas Papernot, Michael Backes, and
  Patrick McDaniel.
\newblock On the (statistical) detection of adversarial examples.
\newblock \emph{arXiv preprint arXiv:1702.06280}, 2017.

\bibitem[Guo et~al.(2017)Guo, Rana, Ciss{\'{e}}, and van~der
  Maaten]{DBLP:journals/corr/abs-1711-00117}
Chuan Guo, Mayank Rana, Moustapha Ciss{\'{e}}, and Laurens van~der Maaten.
\newblock Countering adversarial images using input transformations.
\newblock \emph{CoRR}, abs/1711.00117, 2017.
\newblock URL \url{http://arxiv.org/abs/1711.00117}.

\bibitem[He et~al.(2016)He, Zhang, Ren, and Sun]{he2016deep}
Kaiming He, Xiangyu Zhang, Shaoqing Ren, and Jian Sun.
\newblock Deep residual learning for image recognition.
\newblock In \emph{Proceedings of the IEEE conference on computer vision and
  pattern recognition}, pp.\  770--778, 2016.

\bibitem[Hinton et~al.(2012)Hinton, Deng, Yu, Dahl, Mohamed, Jaitly, Senior,
  Vanhoucke, Nguyen, Sainath, et~al.]{hinton2012deep}
Geoffrey Hinton, Li~Deng, Dong Yu, George~E Dahl, Abdel-rahman Mohamed, Navdeep
  Jaitly, Andrew Senior, Vincent Vanhoucke, Patrick Nguyen, Tara~N Sainath,
  et~al.
\newblock Deep neural networks for acoustic modeling in speech recognition: The
  shared views of four research groups.
\newblock \emph{IEEE Signal processing magazine}, 29\penalty0 (6):\penalty0
  82--97, 2012.

\bibitem[Hu et~al.(2017{\natexlab{a}})Hu, Shen, and
  Sun]{DBLP:journals/corr/abs-1709-01507}
Jie Hu, Li~Shen, and Gang Sun.
\newblock Squeeze-and-excitation networks.
\newblock \emph{CoRR}, abs/1709.01507, 2017{\natexlab{a}}.
\newblock URL \url{http://arxiv.org/abs/1709.01507}.

\bibitem[Hu et~al.(2017{\natexlab{b}})Hu, Shen, and Sun]{hu2017squeeze}
Jie Hu, Li~Shen, and Gang Sun.
\newblock Squeeze-and-excitation networks.
\newblock \emph{arXiv preprint arXiv:1709.01507}, 7, 2017{\natexlab{b}}.

\bibitem[Huang et~al.(2017)Huang, Liu, Van Der~Maaten, and
  Weinberger]{huang2017densely}
Gao Huang, Zhuang Liu, Laurens Van Der~Maaten, and Kilian~Q Weinberger.
\newblock Densely connected convolutional networks.
\newblock In \emph{Proceedings of the IEEE conference on computer vision and
  pattern recognition}, pp.\  4700--4708, 2017.

\bibitem[Huang et~al.(2015)Huang, Xu, Schuurmans, and
  Szepesv{\'a}ri]{huang2015learning}
Ruitong Huang, Bing Xu, Dale Schuurmans, and Csaba Szepesv{\'a}ri.
\newblock Learning with a strong adversary.
\newblock \emph{arXiv preprint arXiv:1511.03034}, 2015.

\bibitem[Katz et~al.(2017)Katz, Barrett, Dill, Julian, and
  Kochenderfer]{katz2017reluplex}
Guy Katz, Clark Barrett, David~L Dill, Kyle Julian, and Mykel~J Kochenderfer.
\newblock Reluplex: An efficient smt solver for verifying deep neural networks.
\newblock In \emph{International Conference on Computer Aided Verification},
  pp.\  97--117. Springer, 2017.

\bibitem[Kurakin et~al.(2016)Kurakin, Goodfellow, and
  Bengio]{kurakin2016adversarial}
Alexey Kurakin, Ian Goodfellow, and Samy Bengio.
\newblock Adversarial machine learning at scale.
\newblock \emph{arXiv preprint arXiv:1611.01236}, 2016.

\bibitem[LeCun et~al.(1998)LeCun, Bottou, Bengio, and
  Haffner]{lecun1998gradient}
Yann LeCun, L{\'e}on Bottou, Yoshua Bengio, and Patrick Haffner.
\newblock Gradient-based learning applied to document recognition.
\newblock \emph{Proceedings of the IEEE}, 86\penalty0 (11):\penalty0
  2278--2324, 1998.

\bibitem[LeCun et~al.(2015)LeCun, Bengio, and Hinton]{lecun2015deep}
Yann LeCun, Yoshua Bengio, and Geoffrey Hinton.
\newblock Deep learning.
\newblock \emph{nature}, 521\penalty0 (7553):\penalty0 436, 2015.

\bibitem[Liu et~al.(2018)Liu, Liu, Su, Cao, and Zhu]{liu2018analyzing}
Mengchen Liu, Shixia Liu, Hang Su, Kelei Cao, and Jun Zhu.
\newblock Analyzing the noise robustness of deep neural networks.
\newblock \emph{arXiv preprint arXiv:1810.03913}, 2018.

\bibitem[Madry et~al.(2017)Madry, Makelov, Schmidt, Tsipras, and
  Vladu]{madry2017towards}
Aleksander Madry, Aleksandar Makelov, Ludwig Schmidt, Dimitris Tsipras, and
  Adrian Vladu.
\newblock Towards deep learning models resistant to adversarial attacks.
\newblock \emph{arXiv preprint arXiv:1706.06083}, 2017.

\bibitem[Meng \& Chen(2017)Meng and Chen]{meng2017magnet}
Dongyu Meng and Hao Chen.
\newblock Magnet: a two-pronged defense against adversarial examples.
\newblock In \emph{Proceedings of the 2017 ACM SIGSAC Conference on Computer
  and Communications Security}, pp.\  135--147. ACM, 2017.

\bibitem[Metzen et~al.(2017)Metzen, Genewein, Fischer, and
  Bischoff]{metzen2017detecting}
Jan~Hendrik Metzen, Tim Genewein, Volker Fischer, and Bastian Bischoff.
\newblock On detecting adversarial perturbations.
\newblock \emph{arXiv preprint arXiv:1702.04267}, 2017.

\bibitem[Papernot et~al.(2016{\natexlab{a}})Papernot, McDaniel, Wu, Jha, and
  Swami]{papernot2016distillation}
Nicolas Papernot, Patrick McDaniel, Xi~Wu, Somesh Jha, and Ananthram Swami.
\newblock Distillation as a defense to adversarial perturbations against deep
  neural networks.
\newblock In \emph{2016 IEEE Symposium on Security and Privacy (SP)}, pp.\
  582--597. IEEE, 2016{\natexlab{a}}.

\bibitem[Papernot et~al.(2016{\natexlab{b}})Papernot, McDaniel, and
  Goodfellow]{DBLP:journals/corr/PapernotMG16}
Nicolas Papernot, Patrick~D. McDaniel, and Ian~J. Goodfellow.
\newblock Transferability in machine learning: from phenomena to black-box
  attacks using adversarial samples.
\newblock \emph{CoRR}, abs/1605.07277, 2016{\natexlab{b}}.
\newblock URL \url{http://arxiv.org/abs/1605.07277}.

\bibitem[Redmon et~al.(2016)Redmon, Divvala, Girshick, and
  Farhadi]{redmon2016you}
Joseph Redmon, Santosh Divvala, Ross Girshick, and Ali Farhadi.
\newblock You only look once: Unified, real-time object detection.
\newblock In \emph{Proceedings of the IEEE conference on computer vision and
  pattern recognition}, pp.\  779--788, 2016.

\bibitem[Ren et~al.(2015)Ren, He, Girshick, and Sun]{ren2015faster}
Shaoqing Ren, Kaiming He, Ross Girshick, and Jian Sun.
\newblock Faster r-cnn: Towards real-time object detection with region proposal
  networks.
\newblock In \emph{Advances in neural information processing systems}, pp.\
  91--99, 2015.

\bibitem[Shafahi et~al.(2018)Shafahi, Huang, Studer, Feizi, and
  Goldstein]{shafahi2018adversarial}
Ali Shafahi, W~Ronny Huang, Christoph Studer, Soheil Feizi, and Tom Goldstein.
\newblock Are adversarial examples inevitable?
\newblock \emph{arXiv preprint arXiv:1809.02104}, 2018.

\bibitem[Simonyan \& Zisserman(2014)Simonyan and Zisserman]{simonyan2014very}
Karen Simonyan and Andrew Zisserman.
\newblock Very deep convolutional networks for large-scale image recognition.
\newblock \emph{arXiv preprint arXiv:1409.1556}, 2014.

\bibitem[Szegedy et~al.(2013)Szegedy, Zaremba, Sutskever, Bruna, Erhan,
  Goodfellow, and Fergus]{DBLP:journals/corr/SzegedyZSBEGF13}
Christian Szegedy, Wojciech Zaremba, Ilya Sutskever, Joan Bruna, Dumitru Erhan,
  Ian~J. Goodfellow, and Rob Fergus.
\newblock Intriguing properties of neural networks.
\newblock \emph{CoRR}, abs/1312.6199, 2013.
\newblock URL \url{http://arxiv.org/abs/1312.6199}.

\bibitem[Vinyals et~al.(2015)Vinyals, Toshev, Bengio, and
  Erhan]{vinyals2015show}
Oriol Vinyals, Alexander Toshev, Samy Bengio, and Dumitru Erhan.
\newblock Show and tell: A neural image caption generator.
\newblock In \emph{Proceedings of the IEEE conference on computer vision and
  pattern recognition}, pp.\  3156--3164, 2015.

\bibitem[Xie et~al.(2017)Xie, Wang, Zhang, Ren, and
  Yuille]{DBLP:journals/corr/abs-1711-01991}
Cihang Xie, Jianyu Wang, Zhishuai Zhang, Zhou Ren, and Alan~L. Yuille.
\newblock Mitigating adversarial effects through randomization.
\newblock \emph{CoRR}, abs/1711.01991, 2017.
\newblock URL \url{http://arxiv.org/abs/1711.01991}.

\bibitem[Xu et~al.(2015)Xu, Ba, Kiros, Cho, Courville, Salakhudinov, Zemel, and
  Bengio]{xu2015show}
Kelvin Xu, Jimmy Ba, Ryan Kiros, Kyunghyun Cho, Aaron Courville, Ruslan
  Salakhudinov, Rich Zemel, and Yoshua Bengio.
\newblock Show, attend and tell: Neural image caption generation with visual
  attention.
\newblock In \emph{International conference on machine learning}, pp.\
  2048--2057, 2015.

\end{thebibliography}
\bibliographystyle{iclr2019_conference}

\newpage
\appendix

\section{Random Shuffle}
\label{appendix:random shuffle}
\begin{figure}[htp]
\centering
\includegraphics[width=1in]{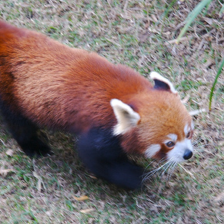}
\includegraphics[width=1in]{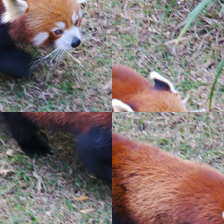}
\includegraphics[width=1in]{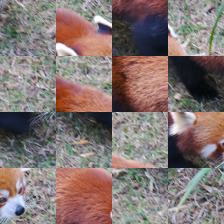}
\includegraphics[width=1in]{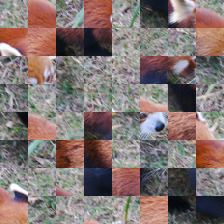}
\caption{An example image that is randomly shuffled after being divided into $1\times 1$, $2\times 2$, $4\times 4$ and $8\times 8$ patches respectively.\label{Random-S}}% All the images used to evaluate sampling from validation set. Before random shuffling, we resize the images to $256 \times 256$ size and then center crop to $224 \times 224$ size. 
\end{figure}
% (I am not quite sure about that. Maybe we need to ask mentor whether we need to put this and how to explain it.)

In this part, we show results of our Random Shuffle experiment. Intuitively, by dropping randomly selected neurons in the neural network, we may let the network learn the relative margins and features better than normal networks. In randomly shuffled images, however, some global patterns of feature distributions are destroyed, so we expect that CNNs with Random Mask would have some trouble extracting feature information and might have worse performance than normal networks. In order to verify our intuition, we compare the test accuracy of a CNN with Random Mask to that of a normal CNN on randomly shuffled images. Specifically speaking, in the experiments, we first train a 0.7-Shallow network along with a normal network on ImageNet dataset. Then we select \(5000\) images from the validation set which are predicted correctly with more than 99\% confidence by both normal and masked networks. We resize these images to $256 \times 256$ and then center crop them to $224 \times 224$. After that, we random shuffle them by dividing them into \(k\times k\) small patches \(k\in\{2,4,8\}\), and randomly rearranging the order of patches. Figure~\ref{Random-S} shows one example of our test images after random shuffling. Finally, we feed these shuffled images to the networks and see their classification accuracy. The results are shown in Table~\ref{shuffle_table}.

\begin{table}[h]
\begin{center}
\begin{tabular}{cccc}
\toprule
\bf Model  & $2\times 2$ &$4\times 4$
& $8\times 8$
\\ \midrule
Normal ResNet-18 &99.58\%	&82.66\%	&17.56\%
\\
0.7-Shallow & 97.36\%	& 64.00\% & 11.94\%
%\\
%Random Mask (adv) &89.??\%	&??.??\% &90.??\%
\\ \bottomrule
\end{tabular}
\end{center}
\caption{The accuracy by using normal and masked networks to classify randomly shuffled test images. \label{shuffle_table}}
\end{table}

From the results, we can see that our network with Random Mask always has lower accuracy than the normal network on these randomly shuffled test images, which indeed accords with our intuition. By randomly shuffling the patches in images, we break the relative positions and margins of the objects and pose negative impact to the network with Random Mask since it may rely on such information to classify. Note that randomly shuffled images are surely difficult for humans to classify, so this experiment might also imply that the network with Random Mask is more similar to human perception than the normal one.

\section{Attack approaches}
\label{appendix:attack approaches}
We first give an overview of how to attack a neural network in some mathematical notations. Let \(\vx\) be the input to the neural network and \(f_\vtheta\) be the function which represents the neural network with parameter \(\vtheta\). The output label of the network to the input can be computed as \(c=\arg\max_if_\vtheta(\vx)\). In order to perform an adversarial attack, we add a small perturbation \(\mathbf{\delta}_x\) to the original image and get an adversarial image \(\vx_{adv}=\vx+\mathbf{\delta}_x\). The new input \(\vx_{adv}\) should look visually similar to the original \(\vx\). Here we use the commonly used $\ell_\infty$-norm metric to measure similarity, i.e., we require that\(||\mathbf{\delta}_x||\le\epsilon\). The attack is considered successful if the predicted label of the perturbed image \(c_{adv}=\arg\max_if_\vtheta(\vx_{adv})\) is different from \(c\).

Generally speaking, there are two types of attack methods: \emph{Targeted Attack}, which aims to change the output label of an image to a specific (and different) one, and \emph{Untargeted Attack}, which only aims to change the output label and does not restrict which specific label the modified example should let the network output.

In this paper, we mainly use the following three attack approaches. \(J\) denotes the loss function of the neural network and \(y\) denotes the true label of \(\vx\).

\begin{itemize}
\item \textbf{Fast Gradient Sign Method (FGSM).} FGSM ~\citep{goodfellow2014} is a one-step untargeted method which generates the adversarial example \(\vx_{adv}\) by adding the sign of the gradients multiplied by a step size \(\epsilon\) to the original benign image \(\vx\). Note that FGSM controls the \(\ell_\infty\)-norm between the adversarial example and the original one by the parameter $\epsilon$.
\[\vx_{adv}=\vx+\epsilon\cdot \text{sign}(\nabla_\vx J(\vx, y)).\]
% \item \textbf{Single-Step Least-Likely Class Method (Step-LL).} Step-LL is similar to FGSM except that Step-LL is a targeted attack method, which chooses the target label to be the least likely class \(y_{LL}\) of \(\vx\). Therefore, Step-ll generates \(\vx_{adv}\) by subtracting the sign of the gradient w.r.t the least likely label.
% \[\vx_{adv}=\vx-\epsilon\cdot sign(\nabla_\vx J(\vx, y_{LL})).\]
\item \textbf{Basic iterative method (PGD).} PGD is a multiple-step attack method which applies FGSM multiple times. To make the adversarial example still stay “close” to the original image, the image is projected to the \(\ell_\infty\)-ball centered at the original image after every step. The radius of the \(\ell_\infty\)-ball is called perturbation scale and is denoted by $\alpha$.
\[\vx_{adv}^0=\vx,\;\;\;\vx_{adv}^{k+1}=Clip_{\vx, \alpha}\left[\vx_{adv}^{k}+\epsilon\cdot \text{sign}(\nabla_{\vx_{adv}^{k}} J(\vx_{adv}^{k}, y))\right].\]
% \item \textbf{Iterative Least-Likely Class Method (I-Step-LL).} I-Step-LL extends the Step-LL to multiple-step attack. It also uses projection after every iteration to avoid the adversarial example being too far away from the original image.
% \[\vx_{adv}^0=\vx,\;\;\;\vx_{adv}^{k+1}=Clip_{\vx, \alpha}\left[\vx_{adv}^{k}-\epsilon\cdot sign(\nabla_{\vx_{adv}^{k}} J(\vx_{adv}^{k}, y_{LL}))\right].\]
\item \textbf{CW Attack.} \cite{DBLP:journals/corr/CarliniW16a} shows that constructing an adversarial example can be formulated as solving the following optimization problem:
\[\vx_{adv}=\argmin_{\vx'} c\cdot g(\vx')+||\vx'-\vx||_2^2,\]
where \(c\cdot g(\vx')\) is the loss function that evaluates the quality of $\vx'$ as an adversarial example and the term \(||\vx'-\vx||_2^2\) controls the scale of the perturbation. More specifically, in the untargeted attack setting, the loss function \(g(\vx)\) can be defined as:
\[g(\vx)=\max\{\max_{i\ne y}\left(f(\vx)_i\right)-f(\vx)_y, -\kappa\},\]
where the parameter $\kappa$ is called confidence.

\end{itemize}

%%\section{Threat Model}
%%According to different knowledge to the model, we can generally define three different kinds of threat model:
%%\begin{itemize}
%%\item \textbf{Black-Box Attack}, which means the threat model have the lowest level of adversary knowledge. Here, we only allow the model to get the final classification results of the inputs without any finer details about the model.
%%\item \textbf{White-Box Attack}, which means the threat model knows everything about the model, including the training data, the algorithm used for training, the structure of the model, the gradient of loss function w.r.t the input image etc.
%%\item \textbf{Grey-Box Attack}, which means the threat model knows something but not all of the model. In our experiment, we define the Grey-Box Attack as the attack method which knows the structure of the model but does not know the gradient information w.r.t the input image. To realize the Grey-Box Attack, similar to ~\citep{DBLP:journals/corr/CarliniW17}, we generate white-box attack on one model with the same (or similar) structure but test the adversarial examples on an independently trained network.
%%\end{itemize}
%%We use all of these three threat models to test the robustness and obtain the interesting properties the neural network structure.

\section{Network Architectures}
\label{appendix: network architecture}
Here we briefly introduce the network architectures used in our experiments. Generally, we apply Random Mask at the shallow layers of the networks and we have tried five different architectures, namely \textbf{ResNet-18}, \textbf{ResNet-50}, \textbf{DenseNet-121} \textbf{SENet-18} and \textbf{VGG-19}. We next illustrate these architectures and show how we apply Random Mask to them.

\subsection{ResNet-18}
\label{ResNet18}
\begin{figure}[h]
\centering
\includegraphics[width=6in]{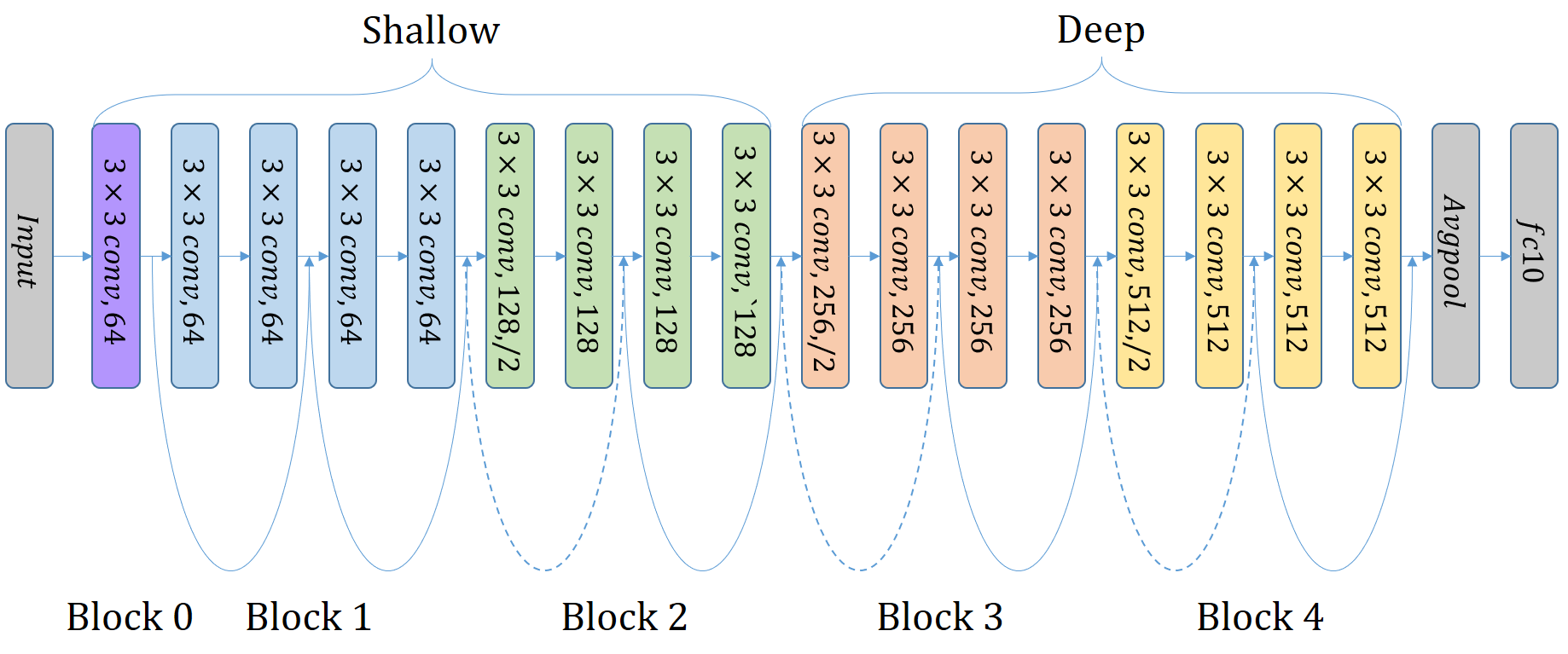}
\caption{The architecture of ResNet-18\label{resnet18}}
\end{figure}

ResNet-18~\citep{he2016deep} contains \(5\) blocks: the $0^{\text{th}}$ block is one single \(3\times 3\) convolutional layer, and each of the rest contains four \(3\times 3\) convolutional layers. Figure~\ref{resnet18} shows the whole structure of ResNet-18. In our experiment, applying Random Mask to a block means applying Random Mask to every layer in it.

\subsection{ResNet-50}

Similar to ResNet-18, ResNet-50~\citep{he2016deep} contains \(5\) blocks and each block contains several \(1\times 1\) and \(3\times 3\) convolutional layers (i.e. Bottlenecks). In our experiment, we apply Random Mask to the \(3\times 3\) convolutional layers in the first three “shallow” blocks. The masked layers in the $1^{\text{st}}$ block are marked by the red arrows.

\begin{figure}[h]
\centering
\includegraphics[width=5in]{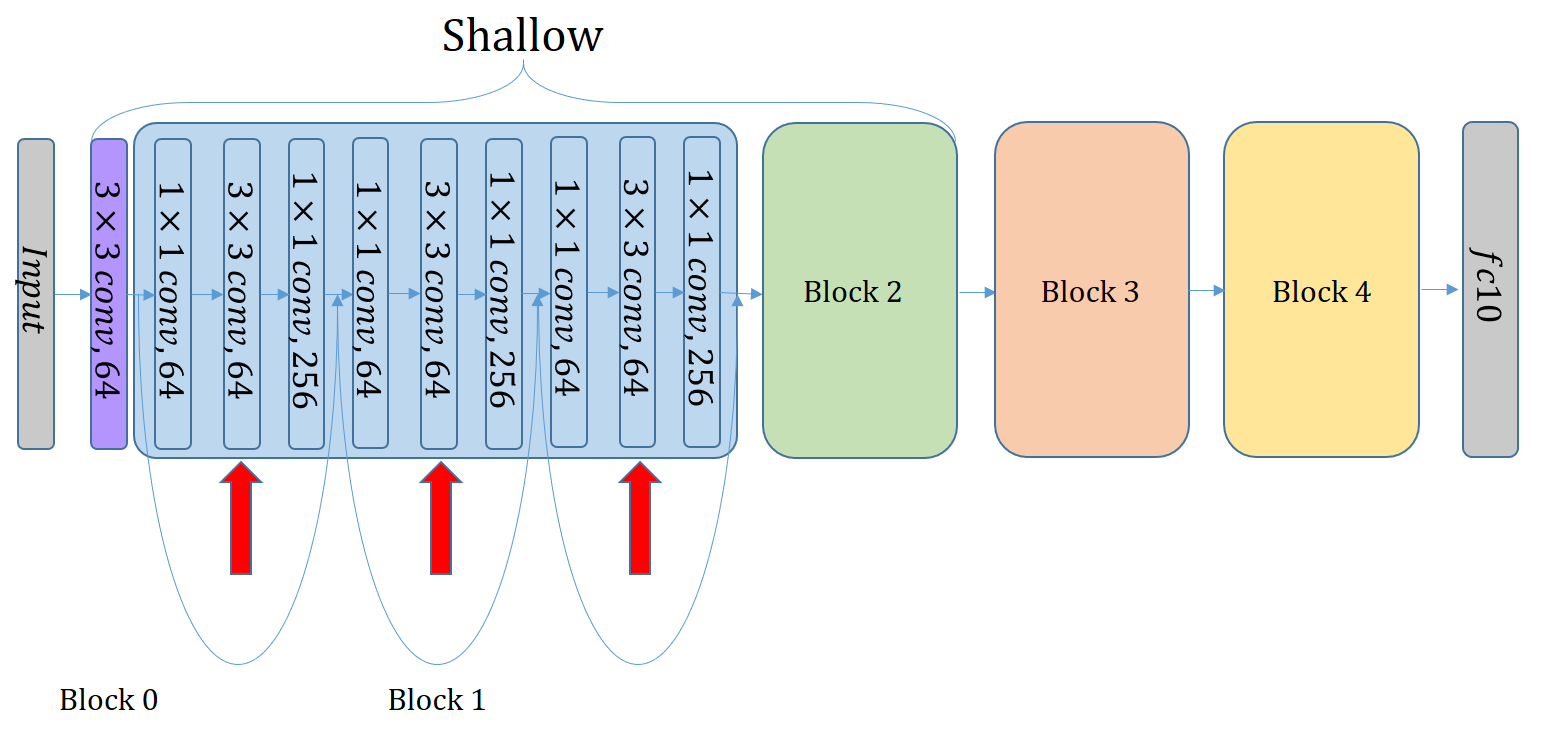}
\caption{The architecture of ResNet-50}
\end{figure}

\subsection{DenseNet-121}

DenseNet-121~\citep{huang2017densely} is another popular network architecture in deep learning researches. It contains \(5\) Dense-Blocks, each of which contains several \(1\times 1\) and \(3\times 3\) convolutional layers. Similar to what we do for ResNet-50, we apply Random Mask to the \(3\times 3\) convolutional layers in the first three ``shallow'' blocks. The growth rate is set to 32 in our experiments.

\begin{figure}[h]
\centering
\includegraphics[width=5in]{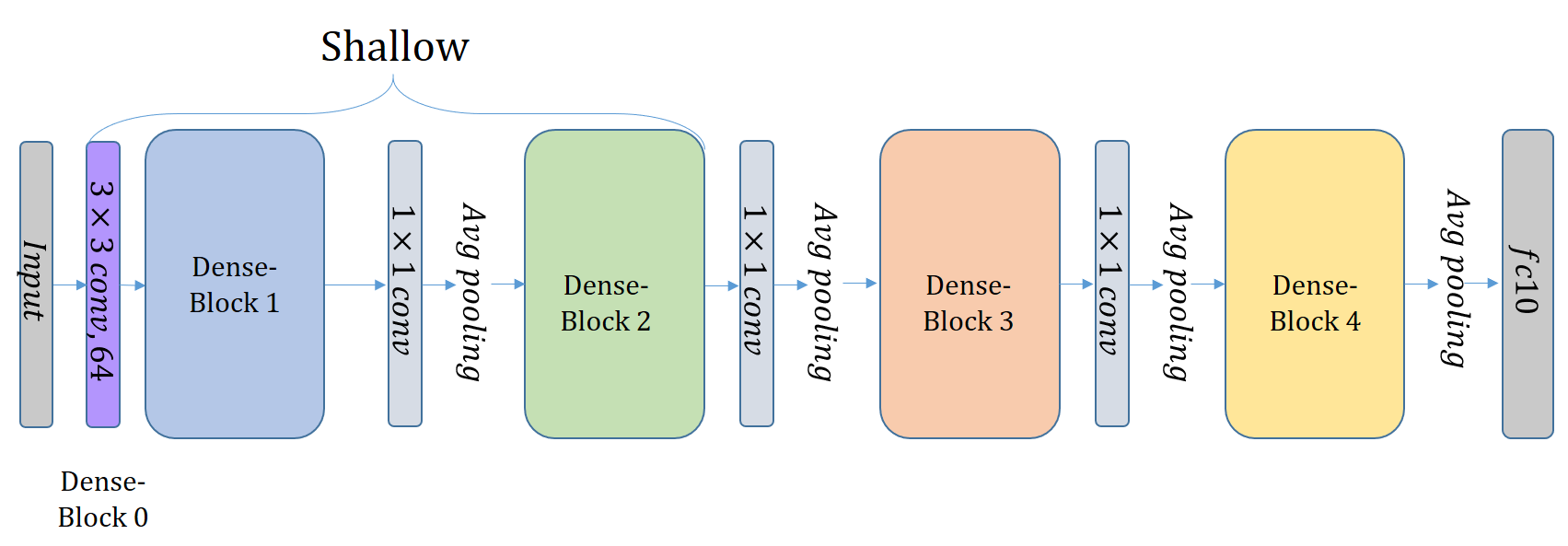}
\caption{The architecture of DenseNet}
\end{figure}

\subsection{SENet}

\begin{figure}[h]
\centering
\includegraphics[width=5in]{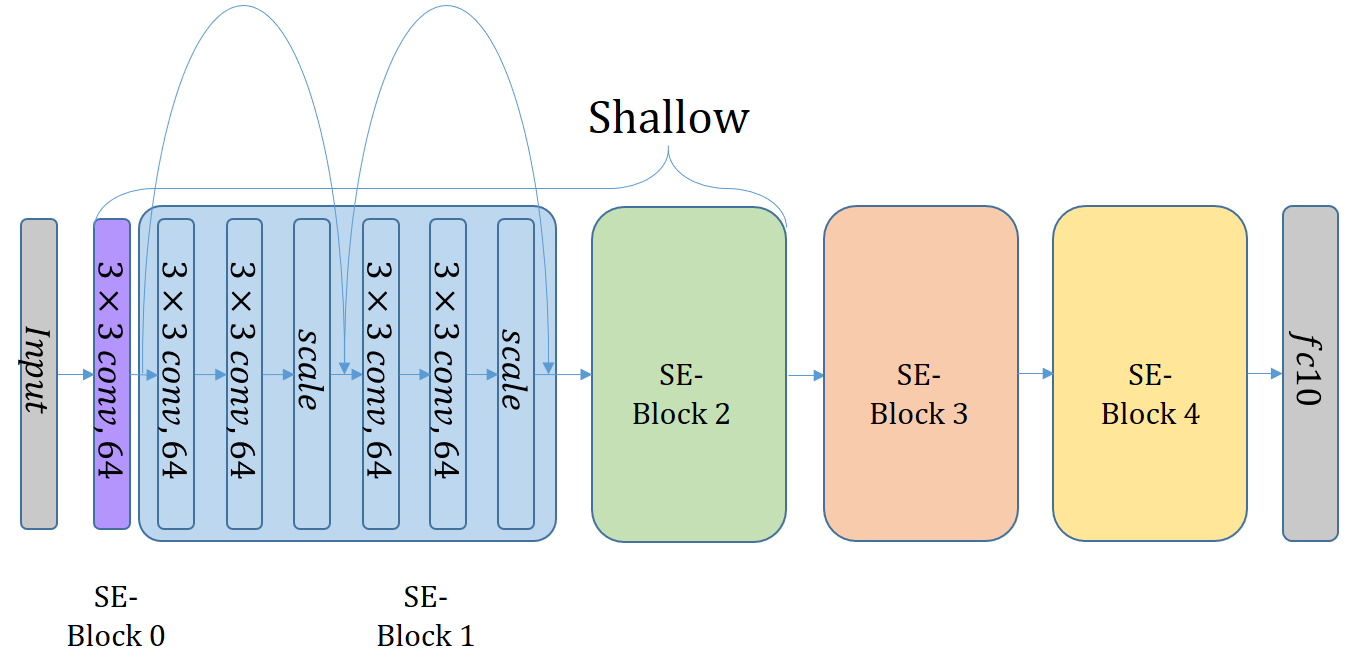}
\caption{The architecture of SENet\label{fig:senet}}
\end{figure}

SENet~\citep{DBLP:journals/corr/abs-1709-01507}, a network architecture which won the first place in ImageNet contest \(2017\), is shown in Figure~\ref{fig:senet}. Note that here we use the pre-activation shortcut version of SENet and we apply Random Mask to the convolutional layers in the first \(3\) SE-blocks.

\subsection{VGG-19}

VGG-19~\citep{simonyan2014very} is a typical neural network architecture with sixteen \(3\times 3\) convolutional layers and three fully-connected layers. We slightly modified the architecture by replacing the final \(3\) fully connected layers with \(1\) fully connected layer as is suggested by recent architectures. We apply Random Mask on the first four \(3\times 3\) convolutional layers.

\begin{figure}[h]
\centering
\includegraphics[width=5in]{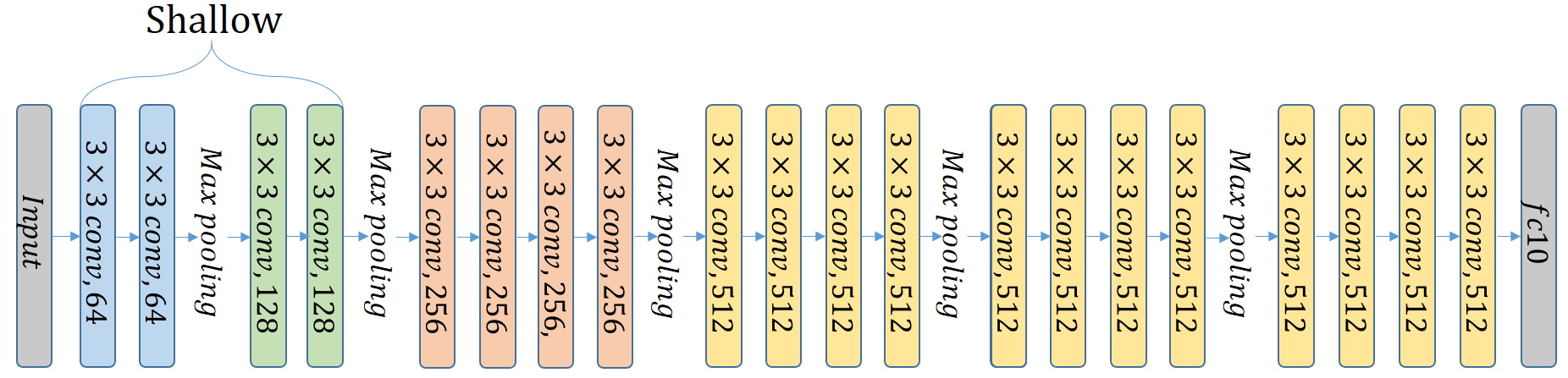}
\caption{The architecture of VGG-19}
\end{figure}

\section{Training Process on CIFAR-10 and MNIST}
\begin{figure}[h]
\centering
\subfigure[CIFAR-10]{
\includegraphics[width=2.0in]{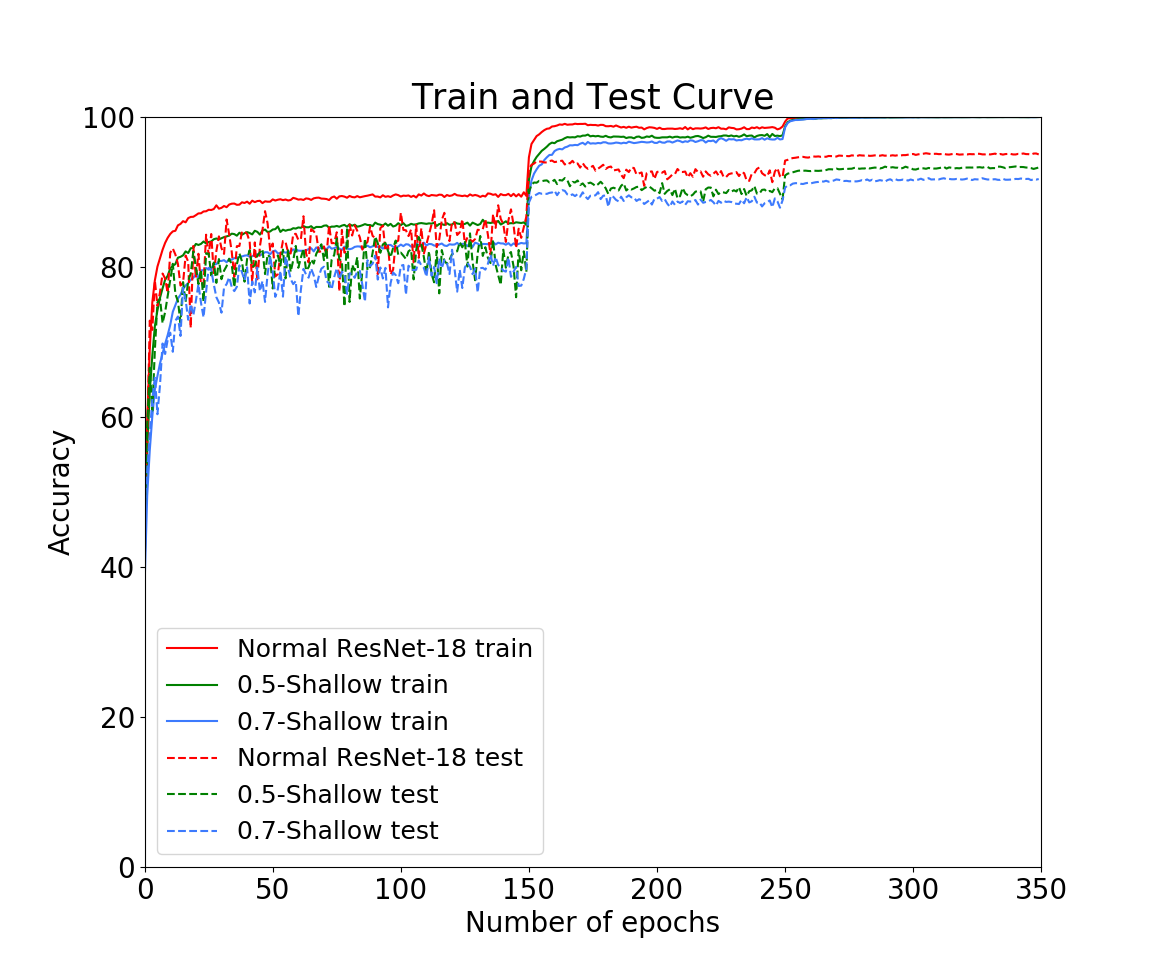}
}
\subfigure[MNIST]{
\includegraphics[width=2.0in]{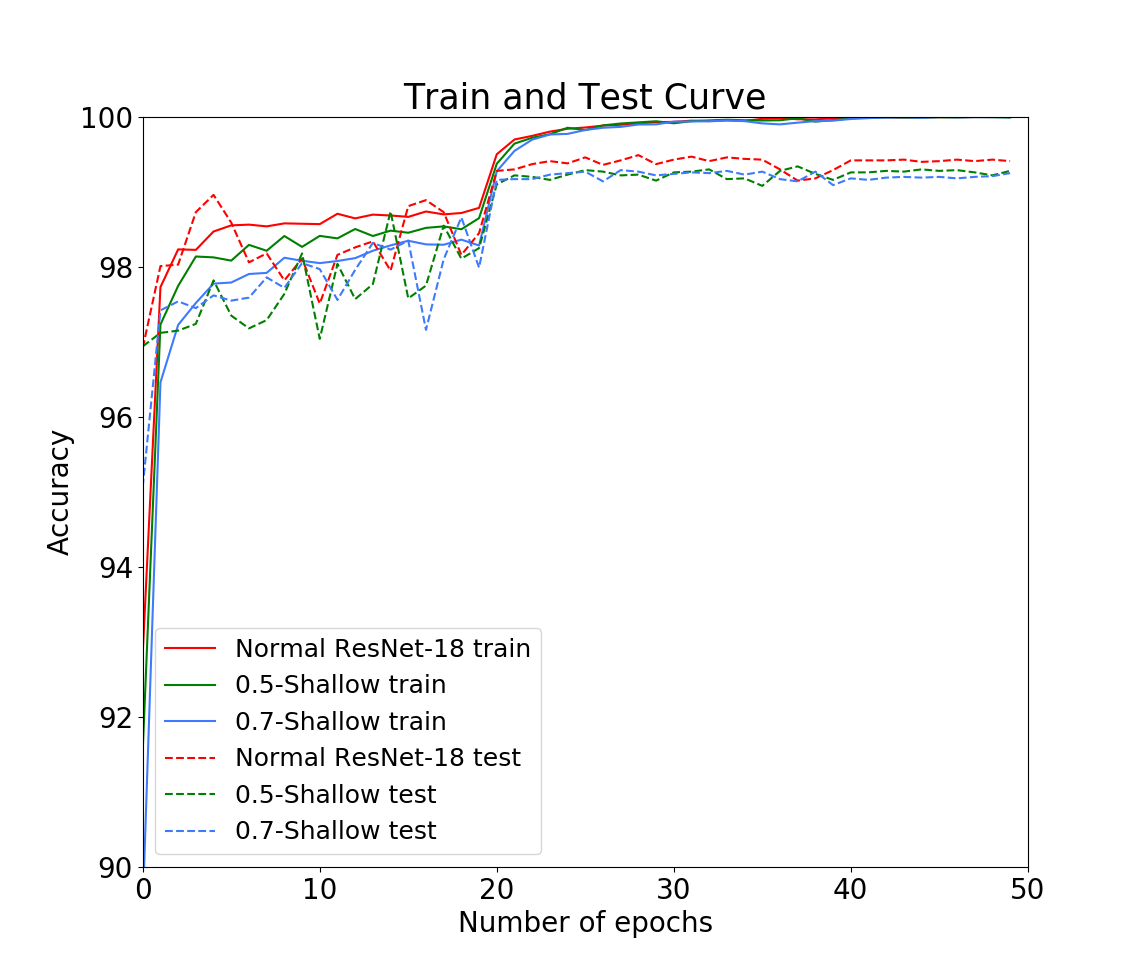}
}
\caption{Train and test curve of normal ResNet-18 and Random Masked ResNet-18 on CIFAR-10 and MNIST. \label{curve}}
\end{figure}

To guarantee our experiments are reproducible, here we present more details on the training process in our experiments. When training models on CIFAR-10, we first subtract per-pixel mean. Then we apply a zero-padding of width \(4\), a random horizontal flip and a random crop of size \(32\times 32\) on train data. No other data augmentation method is used. We apply SGD with momentum parameter \(0.9\), weight decay parameter \(5\times 10^{-4}\) and mini-batch size \(128\) to train on the data for \(350\) epochs. The learning rate starts from \(0.1\) and is divided by 10 when the number of epochs reaches \(150\) and \(250\). When training models on MNIST, we first subtract per-pixel mean. Then we apply random horizontal flip on train data. We apply SGD with momentum parameter \(0.9\), weight decay parameter \(5\times 10^{-4}\) and mini-batch size \(128\) to train on the data for \(50\) epochs. The learning rate starts from \(0.1\) and is divided by 10 when the number of epochs reaches \(20\) and \(40\). Figure~\ref{curve} shows the train and test curves of a normal ResNet-18 and a Random Masked ResNet-18 on CIFAR-10 and MNIST. Different network structures share similar tendency in terms of the train and test curves.

% \begin{figure}[h]
% \centering
% \subfigure[ResNet-18]{
% \includegraphics[width=3.2 in]{./img/TrainCurve_NewNewNew.png}
% }
% \subfigure[ResNet-50]{
% \includegraphics[width=2.6in]{./img/TrainCurve_NewNewNew.png}
% }
% \subfigure[DenseNet-121]{
% \includegraphics[width=2.6in]{./img/TrainCurve_NewNewNew.png}
% }
% \subfigure[SENet-18]{
% \includegraphics[width=2.6in]{./img/TrainCurve_NewNewNew.png}
% }
% \subfigure[VGG19]{
% \includegraphics[width=2.6in]{./img/TrainCurve_NewNewNew.png}
% }
% %\caption{Train and Test Curve\label{curve}}
% \end{figure}

\section{Adversarial Examples Generated by Applying Random Mask}
\subsection{Adversarial Examples that Can “Fool” Human}
\label{appendix: original images}

\begin{figure}[htp]
\centering
\subfigure[Dog]{
\begin{minipage}[b]{0.4in}
\includegraphics[width=0.4in]{./img/9_0_5.png}
\end{minipage}
}
\subfigure[Bird]{
\begin{minipage}[b]{0.4in}
\includegraphics[width=0.4in]{./img/2391_0_2.png}
\end{minipage}
}
\subfigure[Frog]{
\begin{minipage}[b]{0.4in}
\includegraphics[width=0.4in]{./img/451_9_6.png}
\end{minipage}
}
\subfigure[Dog]{
\begin{minipage}[b]{0.4in}
\includegraphics[width=0.4in]{./img/4997_7_5.png}
\end{minipage}
}
\subfigure[Automobile]{
\begin{minipage}[b]{0.4in}
\includegraphics[width=0.4in]{./img/491_9_1.png}
\end{minipage}
}
\subfigure[Ship]{
\begin{minipage}[b]{0.4in}
\includegraphics[width=0.4in]{./img/1724_1_8.png}
\end{minipage}
}
\subfigure[Dog]{
\begin{minipage}[b]{0.4in}
\includegraphics[width=0.4in]{./img/1099_2_5.png}
\end{minipage}
}
\subfigure[Ship]{
\begin{minipage}[b]{0.4in}
\includegraphics[width=0.4in]{./img/2982_1_8.png}
\end{minipage}
}
\subfigure[Bird]{
\begin{minipage}[b]{0.4in}
\includegraphics[width=0.4in]{./img/738_8_2.png}
\end{minipage}
}
\subfigure[Frog]{
\begin{minipage}[b]{0.4in}
\includegraphics[width=0.4in]{./img/452_2_6.png}
\end{minipage}
}
\subfigure[Bird]{
\begin{minipage}[b]{0.4in}
\includegraphics[width=0.4in]{./img/1241_4_2.png}
\end{minipage}
}
\quad
%\subfigure[Airplane]{\includegraphics[width=0.7in]{./img/2117_2_0.png}}
%\subfigure[Dog]{\includegraphics[width=0.7in]{./img/4857_4_5.png}}
\subfigure[Airplane]{
\begin{minipage}[b]{0.4in}
\includegraphics[width=0.4in]{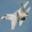}
\end{minipage}
}
\subfigure[Airplane]{
\begin{minipage}[b]{0.4in}
\includegraphics[width=0.4in]{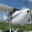}
\end{minipage}
}
\subfigure[Truck]{
\begin{minipage}[b]{0.4in}
\includegraphics[width=0.4in]{./img/451_9.png}
\end{minipage}
}
\subfigure[Horse]{
\begin{minipage}[b]{0.4in}
\includegraphics[width=0.4in]{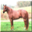}
\end{minipage}
}
\subfigure[Truck]{
\begin{minipage}[b]{0.4in}
\includegraphics[width=0.4in]{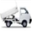}
\end{minipage}
}
\subfigure[Automobile]{
\begin{minipage}[b]{0.4in}
\includegraphics[width=0.4in]{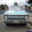}
\end{minipage}
}
\subfigure[Bird]{
\begin{minipage}[b]{0.4in}
\includegraphics[width=0.4in]{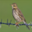}
\end{minipage}
}
\subfigure[Automobile]{
\begin{minipage}[b]{0.4in}
\includegraphics[width=0.4in]{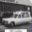}
\end{minipage}
}
\subfigure[Ship]{
\begin{minipage}[b]{0.4in}
\includegraphics[width=0.4in]{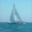}
\end{minipage}
}
\subfigure[Bird]{
\begin{minipage}[b]{0.4in}
\includegraphics[width=0.4in]{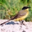}
\end{minipage}
}
\subfigure[Deer]{
\begin{minipage}[b]{0.4in}
\includegraphics[width=0.4in]{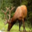}
\end{minipage}
}
\caption{The adversarial examples (upper) shown in Figure~\ref{cmp} along with the original images (lower) from CIFAR-10. \label{cmp_ori}} %ref
\end{figure}

\begin{figure}[htp]
\centering
\subfigure[Mushroom]{
\includegraphics[width=0.7in]{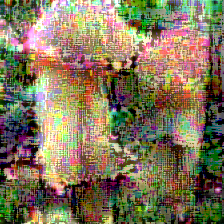}
}
%\subfigure[Dog]{
%\includegraphics[width=0.7in]{./img/Tiny1913_38_36.png}
%}
\subfigure[Monarch Butterfly]{
\includegraphics[width=0.7in]{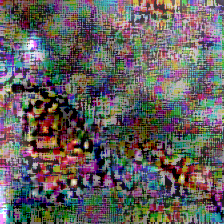}
}
\subfigure[Ladybug]{
\includegraphics[width=0.7in]{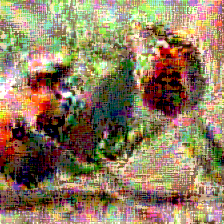}
}
\subfigure[Black Widow]{
\includegraphics[width=0.7in]{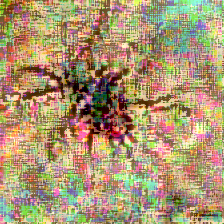}
}
\subfigure[Sulphur Butterfly]{
\includegraphics[width=0.7in]{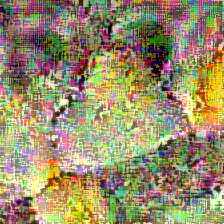}
}
\subfigure[Teddy]{
\includegraphics[width=0.7in]{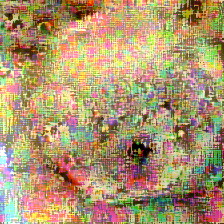}
}
\subfigure[Mushroom]{
\includegraphics[width=0.7in]{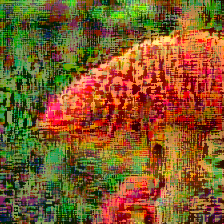}
}
\quad
\subfigure[Tarantula]{
\includegraphics[width=0.7in]{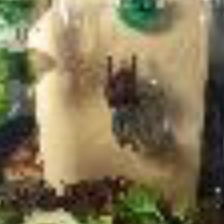}
}
%\subfigure[Dog]{
%\includegraphics[width=0.7in]{./img/Tiny1913_38_36.png}
%}
\subfigure[Black Widow]{
\includegraphics[width=0.7in]{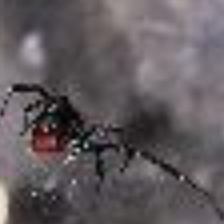}
}
\subfigure[Fly]{
\includegraphics[width=0.7in]{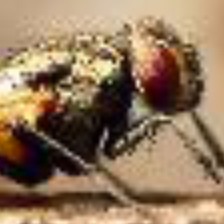}
}
\subfigure[Ladybug]{
\includegraphics[width=0.7in]{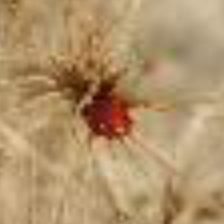}
}
\subfigure[Egyptian Cat]{
\includegraphics[width=0.7in]{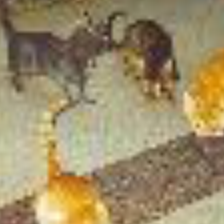}
}
\subfigure[Brain Coral]{
\includegraphics[width=0.7in]{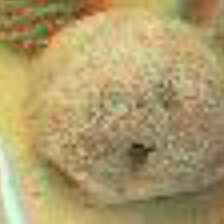}
}
\subfigure[Goldfish]{
\includegraphics[width=0.7in]{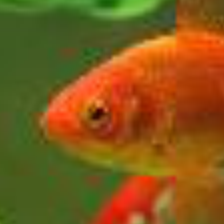}
}
%\subfigure[Airplane]{\includegraphics[width=0.7in]{./img/2117_2_0.png}}
%\subfigure[Dog]{\includegraphics[width=0.7in]{./img/4857_4_5.png}}

\caption{Adversarial examples (upper) generated from Tiny-ImageNet against ResNet-18 with Random Mask of ratio $0.9$ on the $1^{\text{st}},2^{\text{nd}}$ blocks. along with the original images (lower). The attack methods are PGD with scale $64$ and $32$, step size $1$ and step number $40$ and $80$ respectively.\label{cmp_tiny}} %ref
\end{figure}

%The corresponding images is listed on Fig~\ref{cmp_ori}. 
Figure~\ref{cmp_ori} shows some adversarial examples generated from CIFAR-10 along with the corresponding original images. These examples are generated from CIFAR-10 against ResNet-18 with Random Mask of drop ratio $0.8$ on the $0^{\text{th}},1^{\text{st}},2^{\text{nd}}$ blocks and another ResNet-18 with Random Mask of drop ratio $0.9$ on the $1^{\text{st}},2^{\text{nd}}$ blocks. We use attack method PGD with perturbation scale $\alpha=16$ and $\alpha=32$. We also show some adversarial examples generated from Tiny-ImageNet\footnote{\url{https://tiny-imagenet.herokuapp.com/}} along with the corresponding original images in Figure~\ref{cmp_tiny}.

% \subsection{Randomly Selected Adversarial Examples}
% See Figure~\ref{Sample} for a randomly sampled set of images from Tiny-ImageNet along with the corresponding adversarial examples generated against ResNet-18 with Random Mask and normal ResNet-18.

% \label{appendix: random selected}

\section{More Experimental Results}
\label{appendix:More Experiment Result}
\subsection{Black-box Defense under Madry's Setting}
\label{appendixsubsec:Madry Setting}

Here we list the black-box settings in Madry's paper~\citep{madry2017towards}. In their experiments, ResNets are trained by minimizing the following loss:
\[\min_\theta\mathbb{E}_{(x,y)\sim\mathcal{D}}\left[\max_{\delta\in\mathcal{S}}L(\theta,x+\delta,y)\right].\]

The outer minimization is achieved by gradient descent and the inner maximization is achieved by generating PGD adversarial examples with step size \(2\), the number of steps \(7\)  and the perturbation scale \(8\). After training, in their black-box attack setting, they generate adversarial examples from naturally trained neural networks and test them on their models. Both FGSM and PGD adversarial examples have step size or perturbation scale \(8\) and PGD runs for \(7\) gradient descent steps with step size \(2\). 

\begin{figure}[h]
\centering
\includegraphics[width=3.5in]{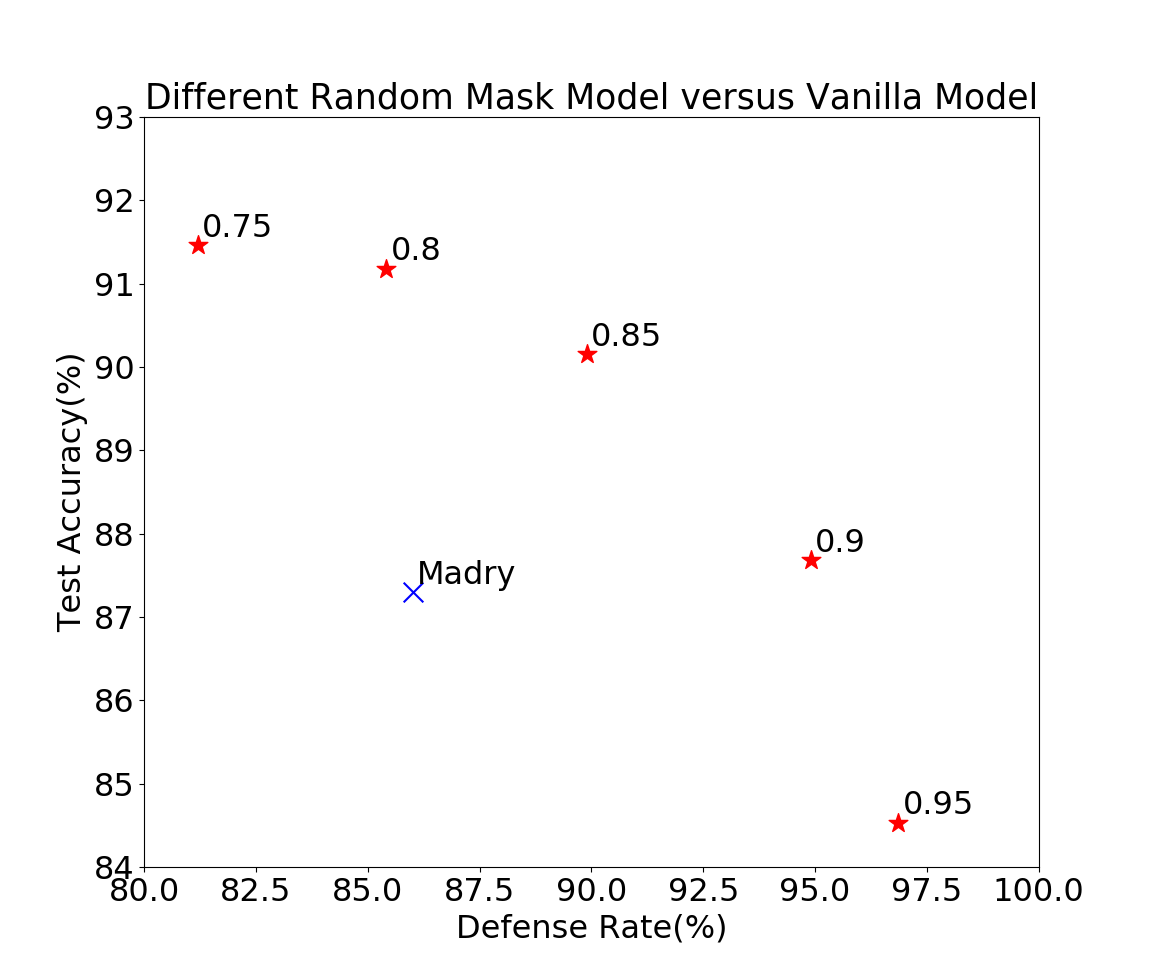}
\caption{Relationship between defense rate against adversarial examples generated by PGD and test accuracy with respect to different drop ratios under Madry's setting~\citep{madry2017towards}. Each red star represents a specific drop ratio with its value written near the star. We can see the trade-off between robustness and generalization.\label{ratio}}
\end{figure}

In Table~\ref{Table 0}, we apply Random Mask to shallow blocks with drop ratio \(0.85\). The ratio is selected by considering the trade-off of robustness and generalization performance, which is shown in Figure~\ref{ratio}. When doing attacks, we generate the adversarial examples in the same way as Madry's paper~\citep{madry2017towards} does.

% \section{Influence of Drop Ratio in Random Mask}
% \label{appendix:ratio}

% The drop ratio is one of the key parameters in Random Mask. Figure~\ref{ratio} shows the results of using different ratios when masking the shallow layers only under Madry setting~\ref{appendixsubsec:Madry Setting}. We can see the trade-off between robustness and generalization.

\newpage
\subsection{Black-box Defense of several network structures on CIFAR-10 and MNIST}
\label{Appendix: several structures}
In this part, we apply Random Mask to five popular network structures - ResNet-18, ResNet-50, DenseNet-121, SENet-18, VGG-19, and test the black-box defense performance on CIFAR-10 and MNIST datasets. 

Since both the intuition (see Section~\ref{sec:Random Mask}) and the extensive experiments (see Section~\ref{subsec:On the Properties of Random Mask} and Appendix~\ref{Appendix: resnet18 experiments}) show that we should apply Random Mask on the relatively shallow layers of the network structure, we would like to do so in this part of experiments. Illustrations of Random Mask applied to these network structures can be found in Appendix~\ref{appendix: network architecture}. In addition, the detailed experiments on ResNet-18 (See Appendix~\ref{Appendix: resnet18 experiments}) show that defense performances are consistent against adversarial examples generated under different settings. Therefore, for brevity, we evaluate the defense performance on adversarial examples generated by PGD only in this subsection.

The results can be found in Table~\ref{Table cifar} and Table~\ref{Table mnist}. Networks in the leftmost column are the target models which defend against adversarial examples. Networks in the first row are the source models to generate adversarial examples by PGD. 0.5-shallow and 0.7-shallow mean applying Random Mask with drop ratio 0.5 and 0.7 to the shallow layers of the network structure whose name lies just above them. The source and target networks are initialized differently if they share the same architecture. All the numbers except the Acc column mean the success rate of defense. The numbers in the Acc column mean the classification accuracy of the target model on clean test data. These results show that Random Mask can consistently improve the black-box defense performance of different network structures.

\begin{table}[h]
\begin{center}
\begin{tabular}{crrrrrc}
\toprule
\multicolumn{1}{c}{\bf Network Structure} 
&\bf $\text{ResNet-18}$
&\bf $\text{ResNet-50}$
&\bf $\text{DenseNet-121}$
&\bf $\text{SENet-18}$
&\bf $\text{VGG-19}$
&\bf Acc\\
\midrule
\multicolumn{1}{c}{Normal ResNet-18} 	&1.42\% & 0.42\% &2.96\%& 1.38\%&9.88\%&95.33\%
\\
0.5-Shallow  &18.46\% & 12.36\%&26.50\%&21.42\%&22.27\%&93.39\%
\\
0.7-Shallow &39.61\% & 33.73\%&47.76\%&44.95\%&39.66\%&91.83\%
\\
\multicolumn{1}{c}{Normal ResNet-50}             & 4.33\%  & 0.10\%  & 3.95\%  & 3.93\%  & 17.58\% & 95.25\% \\
0.5-Shallow & 16.79\% & 5.24\%  & 20.48\% & 19.70\% & 24.82\% & 94.43\% \\
0.7-Shallow & 32.69\% & 19.56\% & 38.34\% & 37.58\% & 37.99\% & 93.46\% \\
\multicolumn{1}{c}{Normal DenseNet-121}             & 4.10\%  & 0.58\%  & 0.60\%  & 3.24\%  & 10.11\% & 95.53\% \\
0.5-Shallow & 12.67\% & 4.18\%  & 10.89\% & 13.30\% & 19.22\% & 93.97\% \\
0.7-Shallow & 23.79\% & 14.54\% & 25.44\% & 27.90\% & 29.61\% & 92.82\% \\
\multicolumn{1}{c}{Normal SENet-18}              & 1.10\%  & 0.52\%  & 2.29\%  & 0.62\%  & 8.02\%  & 95.09\% \\
0.5-Shallow  & 17.47\% & 11.39\% & 23.69\% & 19.62\% & 22.66\% & 93.53\% \\
0.7-Shallow  & 34.19\% & 27.27\% & 43.13\% & 39.05\% & 34.92\% & 92.54\% \\
\multicolumn{1}{c}{Normal VGG-19}               & 6.94\%  & 4.23\%  & 9.98\%  & 6.76\%  & 4.63\%  & 93.93\% \\
0.5-Shallow    & 29.73\% & 25.86\% & 38.66\% & 35.80\% & 26.93\% & 91.73\% \\
0.7-Shallow    & 49.94\% & 45.74\% & 57.74\% & 56.17\% & 46.10\% & 90.11\% \\
 \bottomrule
\end{tabular}
\end{center}
\caption{Black-box experiments on CIFAR-10. Networks in the leftmost column are the target models which defend against adversarial examples. Networks in the first row are the source models to generate adversarial examples by PGD. PGD runs for 20 steps with step size 1 and perturbation scale 16. 0.5-shallow and 0.7-shallow mean applying Random Mask with drop ratio 0.5 and 0.7 to the shallow layers of the network structure whose name lies just above them. All the numbers except the Acc column mean the success rate of defense.\label{Table cifar}}
\end{table}

\begin{table}[h]
\begin{center}
\begin{tabular}{crrrrrc}
\toprule
\multicolumn{1}{c}{\bf Network Structure} 
&\bf $\text{ResNet-18}$
&\bf $\text{ResNet-50}$
&\bf $\text{DenseNet-121}$
&\bf $\text{SENet-18}$
&\bf $\text{VGG-19}$
&\bf Acc\\
\midrule
\multicolumn{1}{c}{Normal ResNet-18} &0.06\%  & 13.80\% & 2.34\%  & 0.10\%  & 8.69\%  & 99.49\% \\
0.5-Shallow&3.45\%  & 17.18\% & 7.41\%  & 4.49\%  & 17.01\% & 99.34\% \\
0.7-Shallow&15.28\% & 47.68\% & 19.29\% & 21.05\% & 38.54\% & 99.29\% \\
\multicolumn{1}{c}{Normal ResNet-50}&3.99\%  & 4.91\%  & 5.79\%  & 3.09\%  & 11.00\% & 99.39\% \\
0.5-Shallow&9.78\%  & 11.93\% & 9.90\%  & 9.09\%  & 14.11\% & 99.32\% \\
0.7-Shallow&9.90\%  & 17.13\% & 10.47\% & 9.27\%  & 19.03\% & 99.28\% \\
\multicolumn{1}{c}{Normal ResNet-50}&1.24\%  & 18.10\% & 0.04\%  & 1.83\%  & 10.36\% & 99.48\% \\
0.5-Shallow&3.09\%  & 18.34\% & 1.52\%  & 3.99\%  & 13.19\% & 99.46\% \\
0.7-Shallow&3.59\%  & 36.63\% & 2.18\%  & 5.78\%  & 20.76\% & 99.38\% \\
\multicolumn{1}{c}{Normal SENet-18}&0.52\%  & 14.31\% & 3.17\%  & 0.08\%  & 12.51\% & 99.41\% \\
0.5-Shallow&3.33\%  & 16.85\% & 5.83\%  & 1.73\%  & 13.67\% & 99.35\% \\
0.7-Shallow&8.84\%  & 26.97\% & 9.36\%  & 8.97\%  & 18.96\% & 99.32\% \\
\multicolumn{1}{c}{Normal VGG-19}&4.09\%  & 33.19\% & 6.25\%  & 6.36\%  & 2.45\%  & 99.48\% \\
0.5-Shallow&9.54\%  & 33.97\% & 9.12\%  & 13.90\% & 9.56\%  & 99.37\% \\
0.7-Shallow&21.00\% & 37.68\% & 20.92\% & 26.44\% & 30.10\% & 99.34\% \\
 \bottomrule
\end{tabular}
\end{center}
\caption{Black-box experiments on MNIST. Networks in the leftmost column are the target models which defend against adversarial examples. Networks in the first row are the source models to generate adversarial examples by PGD. PGD runs for 40 steps with step size $0.01 \times 255$ and perturbation scale $0.3 \times 255$. 0.5-shallow and 0.7-shallow mean applying Random Mask with drop ratio 0.5 and 0.7 to the shallow layers of the network structure whose name lies just above them. All the numbers except the Acc column mean the success rate of defense.\label{Table mnist}}
\end{table}

\newpage

\subsection{White-box}
\label{Appendix: white}

See Table~\ref{Table White-box} for the defense performance of ResNet-18 with Random Mask against white-box attacks on CIFAR-10 dataset. All the numbers except the Acc column mean the success rate of defense. The results on other network architectures are similar.

\begin{table}[h]
\begin{center}
\begin{tabular}{crrrrrrc}
\toprule
\multicolumn{1}{c}{\bf Network Structure} 
&\bf $\text{FGSM}_{1}$ 
&\bf $\text{FGSM}_{2}$ 
&\bf $\text{FGSM}_{4}$ 
&\bf $\text{PGD}_{2}$ 
&\bf $\text{PGD}_{4}$ 
&\bf $\text{PGD}_{8}$
&\bf Acc\\
\midrule
\multicolumn{1}{c}{Normal ResNet-18} 	&81.24\% &65.78\% &51.24\% &24.26\% &3.40\% &0.02\% &95.33\%
\\
0.5-Shallow  &85.22\% &68.65\% &52.04\% &42.35\% &9.11\% &0.42\% &93.39\%
\\
0.7-Shallow  &85.70\% &69.69\% &54.51\%  &49.30\% &19.88\% &3.28\% &91.83\%
\\ \bottomrule
\end{tabular}
\end{center}
\caption{White-box defense performance. $\text{FGSM}_1,\text{FGSM}_2,\text{FGSM}_4$ refer to FGSM with step size 1,2,4 respectively. $\text{PGD}_2,\text{PGD}_4,\text{PGD}_8$ refer to PGD with perturbation scale 2,4,8 and step number 4,6,10 respectively. The step size of all PGD are set to 1.  \label{Table White-box}
}
\end{table}

\subsection{Transferability and Gray-box Defense}
%(Don't know whether the explanation is needed)
Here we show the gray-box defense ability of Random Mask and the transferability of the adversarial examples generated against Random Mask on CIFAR-10 dataset. We generate gray-box attacks in the following two ways. One way is to generate adversarial examples against one trained neural network and test those images on a network with the same structure but different initialization. The other way is specific to our Random Mask models. We generate adversarial examples on one trained network with Random Mask and test them on a network with the same drop ratio but different Random Mask. In both of these two ways, the adversarial knows some information on the structure of the network, but does not know the parameters of it. To see the transferability of the generated adversarial examples, we also test them on DenseNet-121 and VGG-19.
\begin{table}[h]
\begin{center}
\begin{tabular}{|c|ccc|}
\hline
\diagbox{Target}{Source} & Normal ResNet-18 & 0.5-Shallow & 0.7-Shallow\\\hline
Normal ResNet-18&13.91\% & 20.90\% & 28.42\%  \\
0.5-Shallow   & 28.83\% & 20.22\% & 23.24\%  \\
$\text{0.5-Shallow}_{\text{DIF}}$ &28.91\% & 19.37\% & 22.11\%   \\
0.7-Shallow   &49.14\% & 31.30\% & 23.26\%  \\
$\text{0.7-Shallow}_{\text{DIF}}$&48.23\% & 31.77\% & 23.54\%  \\
Normal DenseNet-121  &20.23\% & 22.38\% & 28.05\%  \\
Normal VGG-19       &15.83\% & 17.43\% & 20.85\% \\
\hline
\end{tabular}
\end{center}
\caption{Results on gray-box attacks and transferability. We use FGSM with step size 16 to generate the adversarial examples on source networks and test them on target networks. For target networks, Normal ResNet-18, 0.5-Shallow and 0.7-Shallow represent the networks with the same structure as the corresponding source networks but with different initialization values. $\text{0.5-Shallow}_{\text{DIF}}$ and $\text{0.7-Shallow}_{\text{DIF}}$ represent the networks with the same drop ratios as the corresponding source networks but with different random masks. \label{Table Transfer}}
\end{table}

Table~\ref{Table Transfer} shows that Random Mask can also improve the performance under gray-box attacks. In addition, we find that CNNs with Random Mask have similar performance on adversarial examples generated by our two kinds of gray-box attacks. This phenomenon indicates that CNNs with Random Mask of same ratios have similar properties and catch similar information.

\subsection{Full Information on Experiments Mentioned in Section~\ref{subsec:On the Properties of Random Mask}}

\label{Appendix: resnet18 experiments}
In this part, we will show more experimental results on Random Mask using different adversarial examples, different attack methods and different mask settings on ResNet-18. %The networks used to generate adversarial examples including  Resnet-18, Resnet-50, DenseNet-121, SENet18 and VGG-19. For FGSM, we choose different step size $\epsilon=\{16,32\}$. For PGD, we choose the different step size $\alpha=\{1,2,4\}$, different number of steps $T=\{10,20,40\}$ and different perturbation scale $\lambda=\{4,8,16,32\}$. Note here that all perturbations are applied to the pixel values of images. For CW attack, we set the parameter \(\kappa=40\), which is considered as high-confidence attack. As different adversarial examples \emph{show similar performances}.
More specifically, we choose \(5000\) test images from CIFAR-10 which are correctly classified by the original network to generate FGSM and PGD adversarial examples, and \(1000\) test images for CW attack. 

For FGSM, we try step size \(\epsilon\in\{8, 16, 32\}\), namely $\textbf{FGSM}_{8}$, $\textbf{FGSM}_{16}$, $\textbf{FGSM}_{32}$, to generate adversarial examples. For PGD, we have tried more extensive settings. Let \(\{\epsilon,T,\alpha\}\) be the PGD setting with step size \(\epsilon\), the number of steps \(T\) and the perturbation scale \(\alpha\), then we have tried PGD settings \((1,8,4),(2,4,4),(4,2,4),(1,12,8),(2,6,8),(4,3,8),(1,20,16),(2,10,16),(4,5,16),(1,40,32),\) \((2,20,32),(4,10,32)\) to generate PGD adversarial examples. From the experimental results, we observe the following phenomena. First, we find that the larger the perturbation scale is, the stronger the adversarial examples are. Second, for a fixed perturbation scale, the smaller the step size is, the more successful the attack is, as it searches the adversarial examples in a more careful way around the original image. Based on these observation, we only show strong PGD attack results in the Appendix, namely the settings \((1,20,16)\) ($\textbf{PGD}_{16}$), \((2,10,16)\) ($\textbf{PGD}_{2,16}$) and \((1,40,32)\) ($\textbf{PGD}_{32}$). Nonetheless, our models also perform much better on weak PGD attacks. For CW attack, we have also tried different confidence parameters \(\kappa\). However, we find that for large \(\kappa\), the algorithm is hard to find adversarial examples for some neural networks such as VGG because of its logit scale. For smaller \(\kappa\), the adversarial examples have weak transfer ability, which means they can be easily defensed even by normal networks. Therefore, in order to balance these two factors, we choose \(\kappa=40\) ($\textbf{CW}_{40}$) for DenseNet-121, ResNet-50, SENet-18 and \(\kappa=20\) ($\textbf{CW}_{20}$) for ResNet-18 as a good choice to compare our models with normal ones. The step number for choosing the parameter $c$ is set to $30$. 

Note that the noise of FGSM and PGD is considered in the sense of \(\ell_\infty\) norm and the noise of CW is considered in the sense of \(\ell_2\) norm. All adversarial examples used to evaluate can fool the original network.
%we use FGSM and PGD with \(\lambda\in\{8,16,32\}\) and step size \(\alpha\in\{1,2\}\) and CW attack with confidence \(\kappa=40\) to generate adversarial examples from different trained neural networks, including Resnet-18, Resnet-50, DenseNet-121, SENet18 and VGG-19. For PGD, we also limit the number of steps \(T\in\{10,20,40\}\). Note that for both FGSM and PGD, larger \(\lambda\) leads to greater noise scale and more stronger(successful) adversarial attack. In addition, by setting smaller step size for PGD, the attack searches the adversarial examples in a more careful way and thus enjoys a higher attack success rate. 
Table \ref{Densenet},\ref{Res18},\ref{Res50},\ref{Senet} and \ref{Vgg} list our experimental results. %$\sigma$\text{-}Shallow means applying Random Mask to block \(0,1,2\) in ResNet-18 and the drop ratio is \(\sigma\). 
\(\text{DC}\) means we replace Random Mask with a decreased number of channels in the corresponding blocks to achieve the same drop ratio. \(\text{SM}\) means we use the same mask on all the channels in a layer. \(\times n\) means we multiply the number of the channels in the corresponding blocks by \(n\) times. \(\text{EN}\) means we ensemble five models with different masks of the same drop ratio.

\begin{table}[h]
\begin{tabular}{crrrrrrrr}
\toprule
\multicolumn{1}{c}{\bf Network} &\bf $\text{FGSM}_{8}$&\bf $\text{FGSM}_{16}$&\bf $\text{FGSM}_{32}$&\bf$\text{PGD}_{16}$ &\bf $\text{PGD}_{2,16}$&\bf $\text{PGD}_{32}$ &\bf 
$\text{CW}_{40}$
& \multicolumn{1}{c}{\bf Acc}\\
\midrule
Normal ResNet-18                           & 29.78\% & 14.91\% & 11.53\% & 2.96\%  & 3.44\%  & 2.26\%  & 8.23\%  & 95.33\% \\
0.3-Shallow                                & 55.40\% & 23.29\% & 7.73\%  & 14.53\% & 16.00\% & 5.73\%  & 36.95\% & 94.03\% \\
0.5-Shallow                                & 66.87\% & 30.86\% & 6.65\%  & 26.50\% & 28.65\% & 10.33\% & 54.02\% & 93.39\% \\
0.7-Shallow                                & 79.50\% & 48.57\% & 10.51\% & 47.76\% & 49.62\% & 21.39\% & 73.70\% & 91.83\% \\
0.75-Shallow                               & 83.12\% & 59.22\% & 17.16\% & 56.08\% & 58.81\% & 26.87\% & 77.82\% & 91.46\% \\
0.8-Shallow                                & 85.49\% & 63.01\% & 15.57\% & 63.16\% & 65.16\% & 32.86\% & 81.75\% & 91.18\% \\
0.85-Shallow                               & 88.18\% & 65.27\% & 18.33\% & 69.40\% & 71.21\% & 36.12\% & 85.46\% & 90.15\% \\
0.9-Shallow                                & 94.08\% & 79.93\% & 43.70\% & 83.08\% & 83.72\% & 55.02\% & 89.67\% & 87.68\% \\
0.95-Shallow                               & 96.16\% & 87.36\% & 59.05\% & 89.98\% & 90.13\% & 68.25\% & 90.24\% & 84.53\% \\
0.3-Deep                                   & 28.51\% & 14.62\% & 8.78\%  & 1.95\%  & 2.43\%  & 1.30\%  & 7.88\%  & 95.16\% \\
0.5-Deep                                   & 25.01\% & 10.76\% & 10.24\% & 2.57\%  & 3.81\%  & 4.52\%  & 7.19\%  & 94.94\% \\
0.7-Deep                                   & 23.94\% & 11.23\% & 10.48\% & 3.24\%  & 4.07\%  & 2.64\%  & 10.10\% & 94.61\% \\
0.5-Shallow, 0.5-Deep                      & 58.49\% & 26.34\% & 11.08\% & 19.36\% & 20.77\% & 9.07\%  & 43.59\% & 92.39\% \\
0.3-Shallow, 0.3-Deep                      & 51.03\% & 24.15\% & 10.82\% & 12.67\% & 14.06\% & 6.75\%  & 29.65\% & 94.16\% \\
0.3-Shallow, 0.7-Deep                      & 36.16\% & 11.26\% & 9.16\%  & 7.94\%  & 8.00\%  & 5.77\%  & 23.31\% & 93.44\% \\
0.7-Shallow, 0.7-Deep                      & 64.85\% & 27.43\% & 10.09\% & 32.72\% & 33.84\% & 16.26\% & 62.47\% & 89.78\% \\
0.7-Shallow, 0.3-Deep                      & 74.73\% & 40.58\% & 9.12\%  & 42.95\% & 45.32\% & 19.00\% & 68.58\% & 91.23\% \\
$\text{0.5-Shallow}_{\text{DC}}$           & 36.39\% & 12.15\% & 8.24\%  & 4.68\%  & 5.52\%  & 4.05\%  & 12.72\% & 94.97\% \\
$\text{0.7-Shallow}_{\text{DC}}$           & 43.81\% & 17.74\% & 8.32\%  & 7.51\%  & 9.20\%  & 4.52\%  & 19.34\% & 94.23\% \\
$\text{0.9-Shallow}_{\text{DC}}$           & 49.53\% & 19.00\% & 7.23\%  & 19.33\% & 20.88\% & 10.08\% & 44.80\% & 93.27\% \\
$\text{0.5-Shallow}_{\text{SM}}$           & 77.30\% & 48.86\% & 12.50\% & 44.04\% & 46.58\% & 19.81\% & 72.07\% & 92.57\% \\
$\text{0.7-Shallow}_{\text{SM}}$           & 82.59\% & 48.03\% & 12.30\% & 57.62\% & 57.83\% & 24.81\% & 79.55\% & 89.81\% \\
$\text{0.9-Shallow}_{\text{SM}}$           & 67.06\% & 39.40\% & 16.25\% & 50.40\% & 50.20\% & 29.23\% & 65.38\% & 74.28\% \\
$\text{0.5-Shallow}_{\times 2}$            & 51.25\% & 20.78\% & 10.29\% & 12.51\% & 13.89\% & 5.38\%  & 34.00\% & 94.12\% \\
$\text{0.7-Shallow}_{\times 2}$            & 68.82\% & 30.94\% & 7.22\%  & 29.83\% & 31.74\% & 11.44\% & 60.17\% & 93.01\% \\
$\text{0.9-Shallow}_{\times 2}$            & 88.00\% & 68.83\% & 28.55\% & 66.86\% & 69.42\% & 37.51\% & 82.74\% & 90.49\% \\
$\text{0.9-Shallow}_{\times 4}$            & 82.96\% & 59.64\% & 19.44\% & 59.15\% & 60.72\% & 32.29\% & 78.88\% & 90.57\% \\
$\text{Normal ResNet-18}_{\text{EN}}$      & 35.89\% & 16.24\% & 9.92\%  & 2.22\%  & 2.56\%  & 1.46\%  & 8.58\%  & 96.12\% \\
$\text{0.5-Shallow}_{\times 2, \text{EN}}$ & 55.25\% & 19.84\% & 8.36\%  & 11.86\% & 13.44\% & 5.30\%  & 37.37\% & 95.24\% \\
$\text{0.5-Shallow}_{\text{EN}}$           & 69.35\% & 31.38\% & 7.73\%  & 27.58\% & 29.68\% & 9.97\%  & 58.07\% & 94.56\% \\
$\text{0.7-Shallow}_{\text{EN}}$           & 81.98\% & 51.81\% & 8.57\%  & 50.79\% & 53.88\% & 23.28\% & 77.74\% & 93.31\% \\
$\text{0.85-Shallow}_{\text{EN}}$          & 90.13\% & 66.51\% & 17.91\% & 71.38\% & 72.30\% & 38.25\% & 87.37\% & 91.77\% \\
$\text{0.9-Shallow}_{\text{EN}}$           & 95.37\% & 81.95\% & 43.42\% & 85.14\% & 85.79\% & 56.02\% & 91.36\% & 89.45\%
\\
\bottomrule
\end{tabular}
\caption{Extended experimental results of Section~\ref{subsec:On the Properties of Random Mask}. Adversarial examples generated against \emph{DenseNet-121}. The model trained on CIFAR-10 achieves 95.62\% accuracy on test set. $\sigma\text{-Shallow}_{\text{DC}},\sigma\text{-Shallow}_{\text{SM}},\sigma\text{-Shallow}_{\times n}$ and $\sigma\text{-Shallow}_{\text{EN}}$ mean dropping channels with ratio $\sigma$, applying same mask with ratio $\sigma$, increasing channel number to $n$ times with mask ratio $\sigma$ for every channel and ensemble five models with different masks of same ratio $\sigma$ respectively. The entries in the middle seven columns are success rates of defense under different settings.\label{Densenet}}
\end{table}

\begin{table}[h]
\begin{tabular}{crrrrrrrr}
\toprule
\multicolumn{1}{c}{\bf Network} &\bf $\text{FGSM}_{8}$&\bf $\text{FGSM}_{16}$&\bf $\text{FGSM}_{32}$&\bf$\text{PGD}_{16}$ &\bf $\text{PGD}_{2,16}$&\bf $\text{PGD}_{32}$ &\bf 
$\text{CW}_{20}$
& \multicolumn{1}{c}{\bf Acc}\\
\midrule
Normal ResNet-18 & 26.99\% & 13.91\% & 3.57\%  & 1.42\%  & 1.84\%  & 0.96\%  & 2.19\%  & 95.33\% \\
0.3-Shallow                                        & 48.76\% & 21.32\% & 9.54\%  & 8.14\%  & 9.51\%  & 4.02\%  & 38.87\% & 94.03\% \\
0.5-Shallow                                        & 59.66\% & 30.48\% & 11.60\% & 18.46\% & 21.44\% & 7.70\%  & 60.65\% & 93.39\% \\
0.7-Shallow                                        & 74.00\% & 47.11\% & 15.65\% & 39.61\% & 43.17\% & 16.09\% & 79.04\% & 91.83\% \\
0.75-Shallow                                       & 78.37\% & 56.05\% & 21.44\% & 49.14\% & 52.21\% & 20.31\% & 81.59\% & 91.46\% \\
0.8-Shallow                                        & 81.67\% & 59.14\% & 19.60\% & 55.84\% & 59.63\% & 26.61\% & 82.78\% & 91.18\% \\
0.85-Shallow                                       & 86.31\% & 63.16\% & 22.23\% & 64.73\% & 67.03\% & 31.33\% & 86.06\% & 90.15\% \\
0.9-Shallow                                        & 92.89\% & 77.90\% & 45.63\% & 81.70\% & 82.50\% & 54.12\% & 90.29\% & 87.68\% \\
0.95-Shallow                                       & 95.07\% & 85.40\% & 59.91\% & 88.31\% & 89.64\% & 66.52\% & 90.97\% & 84.53\% \\
0.3-Deep  & 25.96\% & 15.46\% & 7.18\%  & 1.18\%  & 1.28\%  & 0.88\%  & 2.66\%  & 95.16\% \\
0.5-Deep  & 25.21\% & 9.21\%  & 1.44\%  & 2.17\%  & 2.63\%  & 2.31\%  & 3.15\%  & 94.94\% \\
0.7-Deep  & 24.36\% & 9.49\%  & 2.60\%  & 2.36\%  & 3.08\%  & 1.31\%  & 6.62\%  & 94.61\% \\
0.5-Shallow, 0.5-Deep                              & 53.46\% & 24.65\% & 7.08\%  & 13.48\% & 15.65\% & 6.99\%  & 49.95\% & 92.39\% \\
0.3-Shallow, 0.3-Deep                              & 43.32\% & 20.55\% & 4.14\%  & 7.31\%  & 9.36\%  & 4.46\%  & 32.92\% & 94.16\% \\
0.3-Shallow, 0.7-Deep                              & 34.09\% & 11.05\% & 1.58\%  & 6.01\%  & 6.77\%  & 4.56\%  & 24.11\% & 93.44\% \\
0.7-Shallow, 0.7-Deep                              & 61.22\% & 28.11\% & 13.78\% & 27.12\% & 30.24\% & 13.51\% & 69.15\% & 89.78\% \\
0.7-Shallow, 0.3-Deep                              & 70.43\% & 39.15\% & 13.94\% & 36.88\% & 39.81\% & 15.45\% & 74.57\% & 91.23\% \\
$\text{0.5-Shallow}_{\text{DC}}$                   & 32.86\% & 13.89\% & 3.71\%  & 1.93\%  & 2.89\%  & 2.19\%  & 6.10\%  & 94.97\% \\
$\text{0.7-Shallow}_{\text{DC}}$                   & 37.96\% & 16.23\% & 5.05\%  & 4.30\%  & 5.96\%  & 2.65\%  & 15.44\% & 94.23\% \\
$\text{0.9-Shallow}_{\text{DC}}$                   & 48.54\% & 19.10\% & 11.37\% & 14.34\% & 16.01\% & 7.04\%  & 50.62\% & 93.27\% \\
$\text{0.5-Shallow}_{\text{SM}}$                   & 73.96\% & 47.63\% & 16.60\% & 36.19\% & 40.86\% & 15.52\% & 73.68\% & 92.57\% \\
$\text{0.7-Shallow}_{\text{SM}}$                   & 80.80\% & 48.37\% & 15.26\% & 53.69\% & 54.78\% & 22.90\% & 82.34\% & 89.81\% \\
$\text{0.9-Shallow}_{\text{SM}}$                   & 69.15\% & 43.55\% & 20.26\% & 50.68\% & 50.80\% & 28.82\% & 71.62\% & 74.28\% \\
$\text{0.5-Shallow}_{\times 2}$                    & 46.50\% & 21.37\% & 6.06\%  & 6.86\%  & 8.65\%  & 3.59\%  & 39.12\% & 94.12\% \\
$\text{0.7-Shallow}_{\times 2}$                    & 63.37\% & 29.90\% & 12.07\% & 20.70\% & 24.08\% & 7.76\%  & 67.02\% & 93.01\% \\
$\text{0.9-Shallow}_{\times 2}$                    & 84.28\% & 64.47\% & 31.90\% & 62.65\% & 65.24\% & 33.06\% & 85.08\% & 90.49\% \\
$\text{0.9-Shallow}_{\times 4}$                    & 78.24\% & 56.31\% & 23.28\% & 52.38\% & 55.63\% & 25.91\% & 82.26\% & 90.57\% \\
$\text{Normal ResNet-18}_{\text{EN}}$              & 29.66\% & 14.37\% & 3.97\%  & 0.98\%  & 1.24\%  & 0.54\%  & 2.00\%  & 96.12\% \\
$\text{0.5-Shallow}_{\times 2, \text{EN}}$         & 49.16\% & 19.81\% & 6.73\%  & 7.02\%  & 8.50\% & 3.46\%  & 38.12\% & 95.24 \% \\
$\text{0.5-Shallow}_{\text{EN}}$                   & 63.38\% & 30.25\% & 11.05\% & 18.15\% & 21.14\% & 6.71\%  & 63.90\% & 94.56\% \\
$\text{0.7-Shallow}_{\text{EN}}$                   & 77.25\% & 50.07\% & 13.80\% & 41.59\% & 44.78\% & 15.85\% & 80.86\% & 93.31\% \\
$\text{0.85-Shallow}_{\text{EN}}$                  & 88.56\% & 65.23\% & 22.50\% & 65.68\% & 68.31\% & 32.77\% & 88.26\% & 91.77\% \\
$\text{0.9-Shallow}_{\text{EN}}$                   & 94.31\% & 79.47\% & 44.67\% & 82.97\% & 84.05\% & 54.46\% & 90.52\% & 89.4 \%
\\
\bottomrule
\end{tabular}
\caption{Extended experimental results of Section~\ref{subsec:On the Properties of Random Mask}. Adversarial examples are generated against \emph{ResNet-18}. The model trained on CIFAR-10 achieves 95.27\% accuracy on test set. $\sigma\text{-Shallow}_{\text{DC}},\sigma\text{-Shallow}_{\text{SM}},\sigma\text{-Shallow}_{\times n}$ and $\sigma\text{-Shallow}_{\text{EN}}$ mean dropping channels with ratio $\sigma$, applying same mask with ratio $\sigma$, increasing channel number to $n$ times with mask ratio $\sigma$ for every channel and ensemble five models with different masks of same ratio $\sigma$ respectively. The entries in the middle seven columns are success rates of defense under different settings.\label{Res18}}
\end{table}

\begin{table}[h]
\begin{tabular}{crrrrrrrr}
\toprule
\multicolumn{1}{c}{\bf Network} &\bf $\text{FGSM}_{8}$&\bf $\text{FGSM}_{16}$&\bf $\text{FGSM}_{32}$&\bf$\text{PGD}_{16}$ &\bf $\text{PGD}_{2,16}$&\bf $\text{PGD}_{32}$ &\bf 
$\text{CW}_{40}$
& \multicolumn{1}{c}{\bf Acc}\\
\midrule
Normal ResNet-18                           & 29.33\% & 15.14\% & 3.88\%  & 0.42\%  & 0.96\%  & 0.08\%  & 0.00\%  & 95.33\% \\
0.3-Shallow                                & 45.32\% & 18.89\% & 9.16\%  & 4.36\%  & 5.81\%  & 0.92\%  & 1.98\%  & 94.03\% \\
0.5-Shallow                                & 56.26\% & 27.32\% & 10.72\% & 12.36\% & 15.29\% & 3.48\%  & 8.92\%  & 93.39\% \\
0.7-Shallow                                & 70.57\% & 42.40\% & 14.98\% & 33.73\% & 37.56\% & 12.38\% & 33.08\% & 91.83\% \\
0.75-Shallow                               & 77.18\% & 53.01\% & 19.68\% & 42.67\% & 46.72\% & 15.88\% & 39.10\% & 91.46\% \\
0.8-Shallow                                & 80.33\% & 56.21\% & 18.03\% & 52.45\% & 55.54\% & 22.17\% & 47.52\% & 91.18\% \\
0.85-Shallow                               & 84.81\% & 61.02\% & 21.50\% & 62.62\% & 63.80\% & 29.35\% & 53.71\% & 90.15\% \\
0.9-Shallow                                & 92.17\% & 77.68\% & 45.93\% & 81.20\% & 82.50\% & 53.44\% & 66.70\% & 87.68\% \\
0.95-Shallow                               & 94.43\% & 85.54\% & 60.71\% & 88.63\% & 89.16\% & 66.69\% & 71.82\% & 84.53\% \\
0.3-Deep                                   & 27.78\% & 15.03\% & 8.07\%  & 0.42\%  & 0.60\%  & 0.12\%  & 0.00\%  & 95.16\% \\
0.5-Deep                                   & 27.24\% & 10.29\% & 2.47\%  & 0.52\%  & 1.10\%  & 0.50\%  & 0.00\%  & 94.94\% \\
0.7-Deep                                   & 24.81\% & 9.99\%  & 2.50\%  & 0.67\%  & 1.01\%  & 0.20\%  & 0.00\%  & 94.61\% \\
0.5-Shallow, 0.5-Deep                      & 48.78\% & 21.04\% & 6.66\%  & 7.51\%  & 9.79\%  & 2.52\%  & 5.09\%  & 92.39\% \\
0.3-Shallow, 0.3-Deep                      & 42.18\% & 18.20\% & 5.26\%  & 3.96\%  & 5.44\%  & 1.65\%  & 1.64\%  & 94.16\% \\
0.3-Shallow, 0.7-Deep                      & 33.11\% & 11.08\% & 2.27\%  & 2.45\%  & 3.54\%  & 1.29\%  & 0.55\%  & 93.44\% \\
0.7-Shallow, 0.7-Deep                      & 56.39\% & 24.14\% & 12.18\% & 21.86\% & 24.88\% & 9.01\%  & 22.25\% & 89.78\% \\
0.7-Shallow, 0.3-Deep                      & 66.33\% & 36.31\% & 13.09\% & 30.13\% & 33.96\% & 12.09\% & 30.68\% & 91.23\% \\
$\text{0.5-Shallow}_{\text{DC}}$           & 31.56\% & 13.64\% & 4.87\%  & 0.80\%  & 1.30\%  & 0.28\%  & 0.11\%  & 94.97\% \\
$\text{0.7-Shallow}_{\text{DC}}$           & 37.52\% & 15.72\% & 5.38\%  & 1.79\%  & 2.91\%  & 0.56\%  & 0.44\%  & 94.23\% \\
$\text{0.9-Shallow}_{\text{DC}}$           & 44.00\% & 16.90\% & 10.30\% & 8.93\%  & 11.35\% & 3.46\%  & 4.95\%  & 93.27\% \\
$\text{0.5-Shallow}_{\text{SM}}$           & 69.40\% & 41.82\% & 14.27\% & 29.83\% & 33.51\% & 9.60\%  & 26.65\% & 92.57\% \\
$\text{0.7-Shallow}_{\text{SM}}$           & 77.25\% & 44.94\% & 13.80\% & 49.52\% & 51.13\% & 21.24\% & 46.44\% & 89.81\% \\
$\text{0.9-Shallow}_{\text{SM}}$           & 64.32\% & 39.76\% & 19.21\% & 50.16\% & 49.15\% & 28.24\% & 45.18\% & 74.28\% \\
$\text{0.5-Shallow}_{\times 2}$            & 41.51\% & 18.47\% & 6.02\%  & 3.67\%  & 4.80\%  & 0.62\%  & 1.32\%  & 94.12\% \\
$\text{0.7-Shallow}_{\times 2}$            & 58.59\% & 25.92\% & 11.20\% & 14.75\% & 18.29\% & 4.34\%  & 13.77\% & 93.01\% \\
$\text{0.9-Shallow}_{\times 2}$            & 83.05\% & 63.73\% & 29.22\% & 58.85\% & 61.82\% & 28.09\% & 50.11\% & 90.49\% \\
$\text{0.9-Shallow}_{\times 4}$            & 75.74\% & 55.03\% & 21.59\% & 48.93\% & 51.78\% & 20.52\% & 47.08\% & 90.57\% \\
$\text{Normal ResNet-18}_{\text{EN}}$      & 32.70\% & 15.49\% & 4.93\%  & 0.32\%  & 0.84\%  & 0.06\%  & 0.00\%  & 96.12\% \\
$\text{0.5-Shallow}_{\times 2, \text{EN}}$ & 44.90\% & 16.55\% & 6.41\%  & 2.70\%  & 4.00\%  & 0.64\%  & 1.42\%  & 95.24\% \\
$\text{0.5-Shallow}_{\text{EN}}$           & 59.64\% & 26.21\% & 10.17\% & 11.41\% & 14.23\% & 2.48\%  & 8.12\%  & 94.56\% \\
$\text{0.7-Shallow}_{\text{EN}}$           & 73.45\% & 45.60\% & 12.99\% & 33.62\% & 38.49\% & 12.06\% & 32.60\% & 93.31\% \\
$\text{0.85-Shallow}_{\text{EN}}$          & 87.58\% & 62.24\% & 21.81\% & 62.84\% & 64.88\% & 29.10\% & 54.29\% & 91.77\% \\
$\text{0.9-shallow}_{\text{en}}$           & 93.87\% & 79.15\% & 46.44\% & 82.71\% & 83.32\% & 53.87\% & 67.71\% & 89.45\%
\\
\bottomrule
\end{tabular}
\caption{Extended experimental results of Section~\ref{subsec:On the Properties of Random Mask}. Adversarial examples are generated against \emph{ResNet-50}. The model trained on CIFAR-10 achieves 95.69\% accuracy on test set. $\sigma\text{-Shallow}_{\text{DC}},\sigma\text{-Shallow}_{\text{SM}},\sigma\text{-Shallow}_{\times n}$ and $\sigma\text{-Shallow}_{\text{EN}}$ mean dropping channels with ratio $\sigma$, applying same mask with ratio $\sigma$, increasing channel number to $n$ times with mask ratio $\sigma$ for every channel and ensemble five models with different masks of same ratio $\sigma$ respectively. The entries in the middle seven columns are success rates of defense under different settings.\label{Res50}}
\end{table}

\begin{table}[h]
\begin{tabular}{crrrrrrrr}
\toprule
\multicolumn{1}{c}{\bf Network} &\bf $\text{FGSM}_{8}$&\bf $\text{FGSM}_{16}$&\bf $\text{FGSM}_{32}$&\bf$\text{PGD}_{16}$ &\bf $\text{PGD}_{2,16}$&\bf $\text{PGD}_{32}$ &\bf 
$\text{CW}_{40}$
& \multicolumn{1}{c}{\bf Acc}\\
\midrule
Normal ResNet-18                           & 25.53\% & 17.47\% & 8.56\%  & 1.38\%  & 1.78\%  & 0.84\%  & 0.00\%  & 95.33\% \\
0.3-Shallow                                & 46.12\% & 23.30\% & 10.48\% & 9.57\%  & 10.19\% & 4.44\%  & 2.66\%  & 94.03\% \\
0.5-Shallow                                & 57.05\% & 31.01\% & 11.07\% & 21.42\% & 23.17\% & 7.90\%  & 14.61\% & 93.39\% \\
0.7-Shallow                                & 72.67\% & 48.17\% & 15.20\% & 44.95\% & 46.90\% & 19.67\% & 39.89\% & 91.83\% \\
0.75-Shallow                               & 78.23\% & 58.19\% & 21.20\% & 53.56\% & 55.32\% & 23.24\% & 47.33\% & 91.46\% \\
0.8-Shallow                                & 82.27\% & 61.61\% & 19.70\% & 61.55\% & 63.83\% & 31.05\% & 51.14\% & 91.18\% \\
0.85-Shallow                               & 85.80\% & 65.92\% & 22.73\% & 69.82\% & 70.55\% & 34.95\% & 57.36\% & 90.15\% \\
0.9-Shallow                                & 92.93\% & 79.13\% & 48.34\% & 84.55\% & 84.63\% & 55.58\% & 65.63\% & 87.68\% \\
0.95-Shallow                               & 94.77\% & 87.13\% & 63.36\% & 90.63\% & 90.61\% & 69.07\% & 69.14\% & 84.53\% \\
0.3-Deep                                   & 23.76\% & 16.66\% & 9.52\%  & 1.12\%  & 1.26\%  & 0.66\%  & 0.00\%  & 95.16\% \\
0.5-Deep                                   & 23.01\% & 12.19\% & 6.56\%  & 1.73\%  & 2.05\%  & 2.05\%  & 0.00\%  & 94.94\% \\
0.7-Deep                                   & 22.87\% & 11.61\% & 6.63\%  & 2.12\%  & 2.54\%  & 1.29\%  & 0.19\%  & 94.61\% \\
0.5-Shallow, 0.5-Deep                      & 51.20\% & 25.49\% & 10.43\% & 15.05\% & 17.41\% & 7.43\%  & 12.26\% & 92.39\% \\
0.3-Shallow, 0.3-Deep                      & 42.34\% & 22.07\% & 7.84\%  & 8.47\%  & 9.68\%  & 4.44\%  & 1.70\%  & 94.16\% \\
0.3-Shallow, 0.7-Deep                      & 31.43\% & 12.17\% & 6.34\%  & 6.19\%  & 6.25\%  & 4.82\%  & 1.53\%  & 93.44\% \\
0.7-Shallow, 0.7-Deep                      & 57.26\% & 29.36\% & 13.99\% & 31.68\% & 32.58\% & 15.16\% & 30.00\% & 89.78\% \\
0.7-Shallow, 0.3-Deep                      & 68.66\% & 41.32\% & 13.72\% & 40.56\% & 42.30\% & 17.58\% & 35.71\% & 91.23\% \\
$\text{0.5-Shallow}_{\text{DC}}$           & 30.81\% & 14.77\% & 6.08\%  & 2.13\%  & 2.61\%  & 1.83\%  & 0.00\%  & 94.97\% \\
$\text{0.7-Shallow}_{\text{DC}}$           & 34.57\% & 17.32\% & 8.04\%  & 3.65\%  & 4.54\%  & 2.07\%  & 0.19\%  & 94.23\% \\
$\text{0.9-Shallow}_{\text{DC}}$           & 43.46\% & 17.61\% & 10.54\% & 15.15\% & 15.51\% & 7.12\%  & 7.41\%  & 93.27\% \\
$\text{0.5-Shallow}_{\text{SM}}$           & 71.27\% & 49.21\% & 16.27\% & 41.34\% & 43.00\% & 16.93\% & 34.92\% & 92.57\% \\
$\text{0.7-Shallow}_{\text{SM}}$           & 79.48\% & 49.66\% & 15.65\% & 58.96\% & 60.00\% & 26.78\% & 48.28\% & 89.81\% \\
$\text{0.9-Shallow}_{\text{SM}}$           & 65.85\% & 42.59\% & 21.87\% & 52.63\% & 52.91\% & 30.38\% & 43.22\% & 74.28\% \\
$\text{0.5-Shallow}_{\times 2}$            & 44.13\% & 21.71\% & 9.49\%  & 8.91\%  & 9.61\%  & 3.47\%  & 2.65\%  & 94.12\% \\
$\text{0.7-Shallow}_{\times 2}$            & 60.51\% & 30.89\% & 11.58\% & 24.82\% & 26.79\% & 8.36\%  & 21.52\% & 93.01\% \\
$\text{0.9-Shallow}_{\times 2}$            & 85.26\% & 67.91\% & 32.51\% & 67.34\% & 69.14\% & 36.78\% & 52.96\% & 90.49\% \\
$\text{0.9-Shallow}_{\times 4}$            & 78.65\% & 60.05\% & 24.45\% & 57.88\% & 59.63\% & 29.79\% & 50.19\% & 90.57\% \\
$\text{Normal ResNet-18}_{\text{EN}}$      & 27.87\% & 17.80\% & 8.81\%  & 1.02\%  & 1.24\%  & 0.52\%  & 0.00\%  & 96.12\% \\
$\text{0.5-Shallow}_{\times 2, \text{EN}}$ & 45.04\% & 20.25\% & 9.69\% & 7.46\%  & 9.02\% & 3.30\%  & 2.84\%  & 95.24\% \\
$\text{0.5-Shallow}_{\text{EN}}$           & 60.42\% & 31.08\% & 10.77\% & 20.98\% & 22.56\% & 7.09\%  & 13.61\% & 94.56\% \\
$\text{0.7-Shallow}_{\text{EN}}$           & 76.08\% & 51.49\% & 13.19\% & 46.98\% & 49.31\% & 18.95\% & 39.51\% & 93.31\% \\
$\text{0.85-Shallow}_{\text{EN}}$          & 88.64\% & 67.26\% & 23.30\% & 71.44\% & 72.42\% & 37.16\% & 56.36\% & 91.77\% \\
$\text{0.9-Shallow}_{\text{EN}}$           & 94.40\% & 81.32\% & 48.52\% & 85.81\% & 86.28\% & 56.14\% & 66.99\% & 89.45\%
\\
\bottomrule
\end{tabular}
\caption{Extended experimental results of Section~\ref{subsec:On the Properties of Random Mask}. Adversarial examples are  generated against \emph{SENet-18}. The model trained on CIFAR-10 achieves 95.15\% accuracy on test set. $\sigma\text{-Shallow}_{\text{DC}},\sigma\text{-Shallow}_{\text{SM}},\sigma\text{-Shallow}_{\times n}$ and $\sigma\text{-Shallow}_{\text{EN}}$ mean dropping channels with ratio $\sigma$, applying same mask with ratio $\sigma$, increasing channel number to $n$ times with mask ratio $\sigma$ for every channel and ensemble five models with different masks of same ratio $\sigma$ respectively. The entries in the middle seven columns are success rates of defense under different settings.\label{Senet}}
\end{table}

\begin{table}[h]
\begin{tabular}{crrrrrrr}
\toprule
\multicolumn{1}{c}{\bf Network} &\bf $\text{FGSM}_{8}$&\bf $\text{FGSM}_{16}$&\bf $\text{FGSM}_{32}$&\bf$\text{PGD}_{16}$ &\bf $\text{PGD}_{2,16}$&\bf $\text{PGD}_{32}$
& \multicolumn{1}{c}{\bf Acc}\\
\midrule
Normal ResNet-18                           & 37.67\% & 20.25\% & 5.40\%  & 9.88\%  & 12.81\% & 6.72\%  & 95.33\% \\
0.3-Shallow                                & 50.06\% & 23.54\% & 9.53\%  & 14.99\% & 18.66\% & 9.28\%  & 94.03\% \\
0.5-Shallow                                & 57.35\% & 30.52\% & 11.13\% & 22.27\% & 26.87\% & 10.85\% & 93.39\% \\
0.7-Shallow                                & 71.75\% & 47.35\% & 15.47\% & 39.66\% & 44.42\% & 15.91\% & 91.83\% \\
0.75-Shallow                               & 76.81\% & 56.69\% & 19.44\% & 47.88\% & 53.76\% & 18.76\% & 91.46\% \\
0.8-Shallow                                & 79.46\% & 61.45\% & 21.36\% & 56.21\% & 61.01\% & 23.16\% & 91.18\% \\
0.85-Shallow                               & 85.51\% & 66.55\% & 25.35\% & 64.09\% & 67.77\% & 27.53\% & 90.15\% \\
0.9-Shallow                                & 92.58\% & 80.68\% & 51.90\% & 81.65\% & 83.90\% & 52.47\% & 87.68\% \\
0.95-Shallow                               & 95.24\% & 87.10\% & 64.22\% & 88.50\% & 90.04\% & 64.64\% & 84.53\% \\
0.3-Deep                                   & 36.72\% & 18.97\% & 9.65\%  & 9.21\%  & 11.48\% & 6.40\%  & 95.16\% \\
0.5-Deep                                   & 35.93\% & 13.80\% & 2.99\%  & 10.24\% & 12.77\% & 8.95\%  & 94.94\% \\
0.7-Deep                                   & 34.05\% & 13.06\% & 4.04\%  & 10.36\% & 12.70\% & 7.46\%  & 94.61\% \\
0.5-Shallow, 0.5-Deep                      & 52.05\% & 24.88\% & 6.74\%  & 17.83\% & 21.75\% & 10.37\% & 92.39\% \\
0.3-Shallow, 0.3-Deep                      & 47.36\% & 22.74\% & 5.04\%  & 14.44\% & 17.82\% & 9.08\%  & 94.16\% \\
0.3-Shallow, 0.7-Deep                      & 40.38\% & 13.50\% & 3.28\%  & 11.38\% & 14.52\% & 8.12\%  & 93.44\% \\
0.7-Shallow, 0.7-Deep                      & 59.19\% & 28.00\% & 12.13\% & 29.88\% & 34.28\% & 14.12\% & 89.78\% \\
0.7-Shallow, 0.3-Deep                      & 67.14\% & 40.57\% & 13.80\% & 37.63\% & 42.54\% & 16.00\% & 91.23\% \\
$\text{0.5-Shallow}_{\text{DC}}$           & 37.37\% & 16.99\% & 6.62\%  & 9.76\%  & 12.35\% & 8.07\%  & 94.97\% \\
$\text{0.7-Shallow}_{\text{DC}}$           & 42.39\% & 19.90\% & 6.74\%  & 11.93\% & 15.39\% & 8.01\%  & 94.23\% \\
$\text{0.9-Shallow}_{\text{DC}}$           & 47.41\% & 21.12\% & 11.43\% & 20.10\% & 23.16\% & 10.82\% & 93.27\% \\
$\text{0.5-Shallow}_{\text{SM}}$           & 69.61\% & 46.57\% & 14.85\% & 37.31\% & 43.96\% & 15.26\% & 92.57\% \\
$\text{0.7-Shallow}_{\text{SM}}$           & 79.69\% & 48.86\% & 13.87\% & 52.11\% & 56.12\% & 20.35\% & 89.81\% \\
$\text{0.9-Shallow}_{\text{SM}}$           & 67.77\% & 44.38\% & 20.74\% & 49.64\% & 51.28\% & 27.31\% & 74.28\% \\
$\text{0.5-Shallow}_{\times 2}$            & 46.93\% & 21.74\% & 7.11\%  & 14.52\% & 18.17\% & 8.09\%  & 94.12\% \\
$\text{0.7-Shallow}_{\times 2}$            & 60.23\% & 29.72\% & 11.07\% & 23.52\% & 28.11\% & 11.56\% & 93.01\% \\
$\text{0.9-Shallow}_{\times 2}$            & 83.32\% & 66.44\% & 33.11\% & 61.73\% & 66.76\% & 30.33\% & 90.49\% \\
$\text{0.9-Shallow}_{\times 4}$            & 77.68\% & 58.73\% & 24.40\% & 52.07\% & 57.72\% & 23.55\% & 90.57\% \\
$\text{Normal ResNet-18}_{\text{EN}}$      & 39.42\% & 19.53\% & 6.69\%  & 8.67\%  & 12.13\% & 6.46\%  & 96.12\% \\
$\text{0.5-Shallow}_{\times 2, \text{EN}}$ & 49.51\% & 19.29\% & 7.16\%  & 13.71\% & 17.73\% & 8.74\%  & 95.24\% \\
$\text{0.5-Shallow}_{\text{EN}}$           & 60.43\% & 31.07\% & 10.50\% & 21.54\% & 27.09\% & 10.99\% & 94.56\% \\
$\text{0.7-Shallow}_{\text{EN}}$           & 74.11\% & 50.89\% & 13.54\% & 41.26\% & 47.68\% & 16.01\% & 93.31\% \\
$\text{0.85-Shallow}_{\text{EN}}$          & 87.48\% & 67.75\% & 25.62\% & 65.68\% & 69.68\% & 29.26\% & 91.77\% \\
$\text{0.9-Shallow}_{\text{EN}}$           & 94.14\% & 82.46\% & 52.59\% & 83.05\% & 85.46\% & 53.42\% & 89.45\%

\\
\bottomrule
\end{tabular}
\caption{Extended experimental results of Section~\ref{subsec:On the Properties of Random Mask}. Adversarial examples are generated against \emph{VGG-19}. The model trained on CIFAR-10 achieves 94.04\% accuracy on test set. $\sigma\text{-Shallow}_{\text{DC}},\sigma\text{-Shallow}_{\text{SM}},\sigma\text{-Shallow}_{\times n}$ and $\sigma\text{-Shallow}_{\text{EN}}$ mean dropping channels with ratio $\sigma$, applying same mask with ratio $\sigma$, increasing channel number to $n$ times with mask ratio $\sigma$ for every channel and ensemble five models with different masks of same ratio $\sigma$ respectively. The entries in the middle six columns are success rates of defense under different settings.\label{Vgg}}
\end{table}

\end{document}